%% file: elsarticle-template-num.tex
\documentclass[times, review, 10pt]{elsarticle}

\usepackage{amssymb}
\usepackage{graphicx}  % Pour l'inclusion d'images
\usepackage{caption}   % Pour personnaliser les légendes
\usepackage{subcaption}
\usepackage{hyperref}
\usepackage{lipsum} 
\usepackage{xcolor}
\usepackage{cleveref}
\usepackage{comment}
\usepackage{makecell}

\newcommand{\Florent}[1]{\textcolor{black}{#1}} 
\newcommand{\Yann}[1]{\textcolor{black}{#1}} 

\journal{Pattern Recognition}

\begin{document}

\begin{frontmatter}

%% Title, authors and addresses

%% use the tnoteref command within \title for footnotes;
%% use the tnotetext command for theassociated footnote;
%% use the fnref command within \author or \address for footnotes;
%% use the fntext command for theassociated footnote;
%% use the corref command within \author for corresponding author footnotes;
%% use the cortext command for theassociated footnote;
%% use the ead command for the email address,
%% and the form \ead[url] for the home page:
%% \title{Title\tnoteref{label1}}
%% \tnotetext[label1]{}
%% \author{Name\corref{cor1}\fnref{label2}}
%% \ead{email address}
%% \ead[url]{home page}
%% \fntext[label2]{}
%% \cortext[cor1]{}
%% \affiliation{organization={},
%%             addressline={},
%%             city={},
%%             postcode={},
%%             state={},
%%             country={}}
%% \fntext[label3]{}

\title{Mixture-of-experts for handwriting trajectory reconstruction from IMU sensors}

%% use optional labels to link authors explicitly to addresses:
%% \author[label1,label2]{}
%% \affiliation[label1]{organization={},
%%             addressline={},
%%             city={},
%%             postcode={},
%%             state={},
%%             country={}}
%%
%% \affiliation[label2]{organization={},
%%             addressline={},
%%             city={},
%%             postcode={},
%%             state={},
%%             country={}}

\author[1]{Florent Imbert}
\ead{florent.imbert@irisa.fr}
\author[1]{Eric Anquetil}
\author[2]{Yann Soullard}
\author[2]{Romain Tavenard}

\affiliation[1]{organization={IRISA, Universite de Rennes, INSA Rennes},%Department and Organization
            city={Rennes},
            country={France}}
            
\affiliation[2]{organization={IRISA, Universite Rennes 2},%Department and Organization
            city={Rennes},
            country={France}}

\begin{abstract}
%% Text of abstract
The use of digital pens for online handwriting trajectory reconstruction is a prevalent method for human-computer interaction. In this study, we focus on a digital pen equipped with sensors where we aim at reconstructing the online handwriting trajectory. This pen enables writing on any surface and preserving the digital trace of handwriting. This type of pen could be used as an aid to learning to write in classroom.
%A previous study has highlighted the need for a specific processing of the hovering parts. 
In this paper, we propose a new approach learning to finely reconstruct the touching trajectories while precisely analyzing the hovering part in order to position the next touching trace correctly. This relies on a Mixture-Of-Experts (MOE) approach. The first expert is dedicated for the pencil touch, and is named touching expert model. The second one is dedicated for the hovering pen trajectory, and is named hovering expert model. We improve on the learning of each of these experts based on additional context or specific examples. 
In addition we introduce a novel public benchmark dataset, to enable future research and comparisons in the field of handwriting reconstruction. The results demonstrates a significant enhancement compared to its primary competitors
\end{abstract}

%%Graphical abstract
\begin{graphicalabstract}
\includegraphics[width=\textwidth]{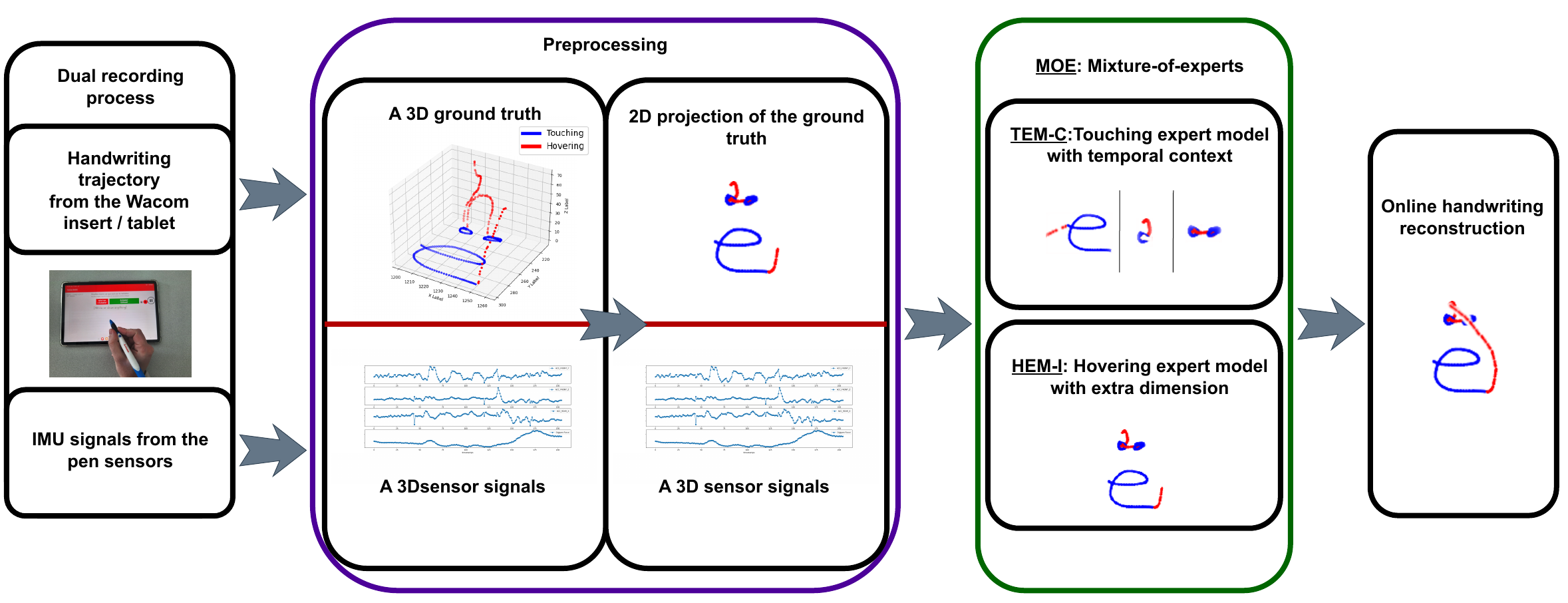}
\end{graphicalabstract}

%%Research highlights
\begin{highlights}
    \item a mixture-of-experts approach made of two expert neural networks, one for the touching trajectory, the other for the hovering parts of the trajectory;
    \item integration of a temporal context reflecting physics and dynamics to improve the expert model dedicated to the touching part, and is also intended to facilitate joining with the other expert model;
    \item taking into account the extra dimension present in the hovering parts linked to the height of the pen to improve the expert model dedicated to the hovering part;
    \item a mixture-of-experts, with each expert optimized both on it's dedicated task and to facilitate collaboration with the other expert, to obtain a mixture of experts forming a coherent and robust system;
    \item a new public database that will serve as a benchmark to help advance research.
\end{highlights}

\begin{keyword}
%% keywords here, in the form: keyword \sep keyword
Online Handwriting \sep Trajectory Reconstruction \sep Digital Pen \sep Inertial Measurement Units \sep Deep Neural Network
%% PACS codes here, in the form: \PACS code \sep code

%% MSC codes here, in the form: \MSC code \sep code
%% or \MSC[2008] code \sep code (2000 is the default)

\end{keyword}

\end{frontmatter}

%% \linenumbers

%% main text
\input{Sections/Introduction}

\input{Sections/Related_works}

\input{Sections/Contributions}

\input{Sections/Results}

\input{Sections/Limitations}

\input{Sections/Conclusion}

\input{Sections/Acknowledgments}

%% The Appendices part is started with the command \appendix;
%% appendix sections are then done as normal sections
%% \appendix

%% \section{}
%% \label{}

%% If you have bibdatabase file and want bibtex to generate the
%% bibitems, please use
%%
%%  \bibliographystyle{elsarticle-num} 
%%  \bibliography{<your bibdatabase>}

%% else use the following coding to input the bibitems directly in the
%% TeX file.

\bibliographystyle{elsarticle-num} 
\bibliography{bib}

%% else use the following coding to input the bibitems directly in the
%% TeX file.

\vspace{0.4cm}

\input{Sections/bio}
\end{document}

%% file: Sections/Introduction.tex
\section{Introduction}
\label{Introduction}

Digital devices play a crucial role in enhancing the learning experience for both students and teachers by facilitating active learning techniques and offering immediate feedback \citep{simonnet2019evaluation}. The literature on e-learning \citep{Atilola_Valentine_Kim_Turner_McTigue_Hammond_Linsey_2014, Barreto, BONNETONBOTTE} highlights the accuracy and reliability of computer-based analysis for generating relevant feedback for correction or guidance. Based on this, pen-based tablet applications have been developed to provide personalized feedback \citep{krichen2022combination, el}.

\begin{figure}[!ht]
\begin{center}
\includegraphics[scale=0.07]{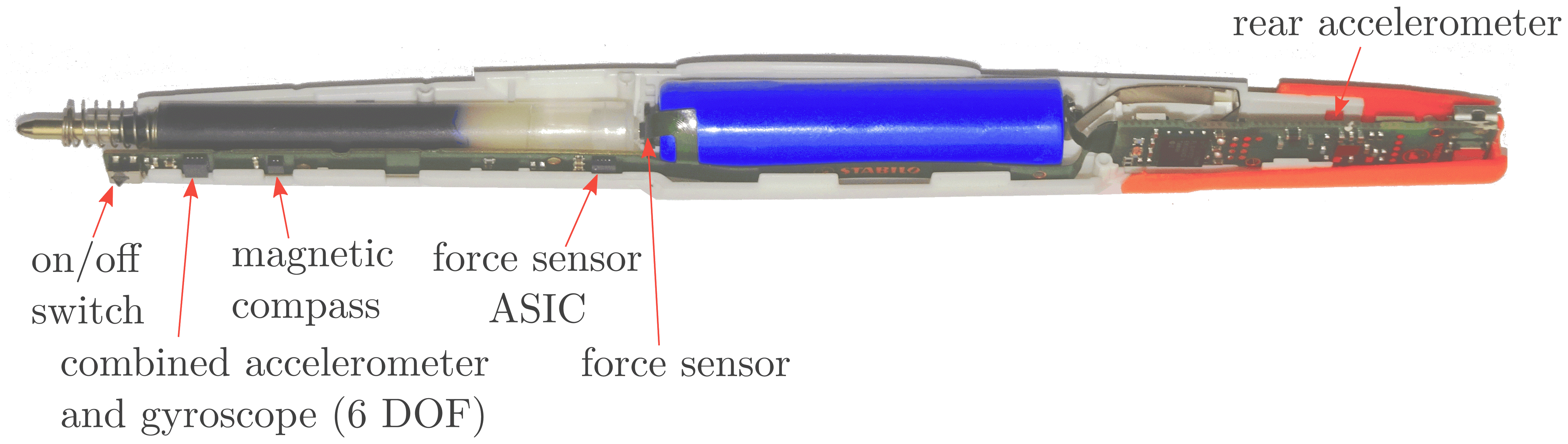}
\caption{© STABILO International Digipen's sensor location }
\label{fig:Sensor_location}
\end{center}
\end{figure}

Despite the increasing reliance on digital platforms, there remains a need for children to learn writing on paper, as it remains the most widely used surface. To address this, digital pens, such as the Digipen stylus developed by STABILO, have been equipped with kinematic sensors (Inertial Measurement Units) to track pen movements (Fig. \ref{fig:Sensor_location}). Such a pen allows to capture handwriting gestures on any surface\Yann{, including paper, and it allows to} visualize and analyze the handwriting reconstruction on a digital medium (e.g. tablet, computer).

This work focuses \Yann{on the challenging task of reconstructing the digital handwriting trajectory from the Digipen stylus.} % using a deep learning-based approach.} Recent advancements in deep neural network architectures have significantly benefited applications in remote sensing and tracking systems.
While Inertial Measurement Units (IMUs) are commonly used in tracking systems due to their cost-effectiveness, \Yann{their inaccurate signals due to high noise levels are problematic. Unlike handwriting recognition tasks, where a label exists for the global shape of the handwriting, trajectory reconstruction involves a precise reconstruction at each time frame by analyzing local displacement. }%The designed system need to extract local displacement features based on a local view to predict positions accurately. 
%The designed system must produce the most accurate trajectory reconstruction possible for e-learning applications, where precise feedback to learners is essential. }%This is particularly %evident in the context of handwriting trajectory reconstruction, where precision is 
%crucial for e-learning applications in order to give appropriate feedback to the learner.}

\Florent{This research hypothesizes that the precise reconstruction of handwriting trajectories from IMU data can significantly improve feedback mechanisms in educational applications. This enhances both learning outcomes and user experience, while preserving the natural experience of writing on paper. By enabling real-time handwriting instruction and correction, this approach offers the dual advantage of writing on physical paper \Yann{and benefiting from digital analysis and feedback, helping to make e-learning tools more effective. Thus, the designed system must produce the most accurate trajectory reconstruction possible for e-learning applications, where precise feedback to learners is essential. }}

Previous studies have utilized IMU sensors for various applications, including recognizing predefined movements \cite{ring,s,math}, reconstructing pedestrian trajectories \cite{Shoes}, and estimating upper limbs kinematics during industrial gestures of pick and place \citep{DIGO}. %\cite{bracelet,ring,s,math}
Several previous works investigated handwriting reconstruction using the Digipen stylus. \cite{wehbi2022surface} propose a Convolutional Neural Network for this task. This extends the works of \cite{ott2022joint} to multiple writers. 
\citep{Swaileh2023} is the third attempt to reconstruct handwriting trajectories from the Digipen using deep neural networks. It has produced promising results in terms of handwriting reconstruction. Their approach is dedicated to the reconstruction of touching parts, which are the only parts for which a reliable ground truth is available during training. 

\Florent{IMU sensors, while useful for tracking motion, are prone to accumulating errors over time, a phenomenon known as sensor drift. \Yann{In the field of handwriting trajectory reconstruction, prior methods \cite{Miyagawa,pan}} have faced significant limitations due to sensor drift. This drift causes imprecision in long-duration or continuous handwriting tasks, making it difficult to reconstruct accurate trajectories. In particular, previous approaches \cite{wehbi2022surface,6975206} that treated handwriting as a single continuous motion without segmenting the writing into distinct parts often amplified the effects of this drift, leading to inaccurate global  reconstructions.}

\Florent{To address this issue, \Yann{we propose a novel handwriting trajectory reconstruction approach consisting of two expert networks to deal with the different nature of IMU signals. Our proposal is based on the key idea of separating the task into two parts: reconstructing the handwriting trajectory when the stylus touches the surface and when the pen is up, which is called the hovering trajectory. } \Yann{In fact, touching trajectories are 2-dimensional trajectories referring to the writing itself and where ground truth can be obtained using double Digipen-Wacom acquisitions (Fig. \ref{fig:rec_process}). Hovering trajectories are 3-dimensional trajectories corresponding to a target trajectory towards the next part of the writing and where ground truth signals can only be recovered under a 7mm height. This is an additional difficulty to take into account when dealing with hovering trajectories and that, to our knowledge, has not been properly considered in the literature. Thus, in this work, the goal is twofold. First, each type of movement (2D touching and 3D hovering parts) has its own dynamic properties, and our approach assigns specialized expert models to handle these segments independently. Second, by separating the two tasks, the proposed approach aims to reduce the cumulative error introduced by sensor drift. This motivates our choice of designing a mixture-of-experts approach with one expert neural network for each of these specific tasks.} }

%The authors also highlight} the importance of dynamics in IMU trajectory reconstruction. 

% By focusing on the 2D touching trajectories when the pen is in contact with the writing surface, and isolating the 3D hovering trajectories when the pen is lifted, we reduce the cumulative error introduced by sensor drift. In addition, each type of movement has its own dynamic properties, and our approach assigns specialized expert networks to handle these segments independently.}
%In this work, the focus is on the challenging task of reconstructing handwriting trajectories. Unlike handwriting recognition tasks, where a label exists for the global shape of the handwriting, trajectory reconstruction involves a precise reconstruction at each time frame. Models need to extract local displacement features based on a local view to predict positions accurately. 

%We present a novel handwriting trajectory reconstruction approach from IMU sensors. 
%The pencil touch 2D trajectory, which is the writing itself, for which we have a 2-dimensional ground truth; and the pencil lift 3D trajectory, which corresponds to a target trajectory towards the next part of the writing. For the 3D pencil lift, the dynamics and ground truth are no longer the same. The ground truth, which this time is in 3 dimensions, can be lost if the user raises the pencil too high. This motivates our choice of having one expert network for each of these specific tasks.
Our main contributions are presented below: 
\begin{itemize}
    \item a mixture-of-experts approach made of two expert neural networks, one for the touching trajectory, the other for the hovering parts of the trajectory;
    \item \Yann{the integration of temporal context into the expert model dedicated to the touching part, reflecting physics and dynamics and facilitating the consistency with the other expert model};
    \item \Yann{Enhancing the capabilities of the expert model dedicated to the hovering part by taking into account the additional dimension due to the height of the pen during training};
    \item a mixture-of-experts, \Yann{where each expert is specialized on its dedicated task and where the collaboration with the other expert is facilitated}, to obtain a mixture of experts forming a coherent and robust system;
    \item a new public database that will serve as a benchmark to help advance research.
\end{itemize}
The training and testing phases are described in details, to enable reproducibility of the experiments relying on an open database.
 
\Florent{The rest of the paper is organized as follows, related works are presented in the following section \ref{Related works}. A short review of the metrics used for handwriting evaluation is presented in section \ref{EHR}. Section \ref{Contributions} introduces our innovative approach to expert neural networks dedicated to each part of writing (touching and hovering parts). We then present improvements for each expert. The experimental results are presented and discussed both quantitatively and qualitatively in section \ref{Results}. An ablation study shows the impact of each contribution. Finally, we conclude and discuss perspectives. }

%% file: Sections/Related_works.tex
\section{Related works}
\label{Related works}

The field of digital devices for note-taking, drawing, and handwriting learning is rapidly expanding. However, most systems use display for digital handwriting acquisition, with only a few incorporating styluses equipped with motion tracking systems to reconstruct handwriting trajectories. In this section, we focus related works on handwriting trajectory reconstruction and the use of Inertial Measurement Unit (IMU) sensors for handwriting recognition.

\subsection{Handwriting Trajectory Reconstruction}

There are various systems dedicated to the acquisition of digital handwriting. Most of them use pen-based tablets with display, such as the Samsung S Pen, Apple Pencil, Microsoft Surface Pen, and Wacom offering precise tracking but dedicated to specific devices. For these systems, the pen is powered by the electromagnetic field generated by the Electro-Magnetic Resonance (EMR) sensor located under the diaplay. Another approach enables writing on paper using pens equipped with tools to capture handwriting gestures. This can involve embedding cameras in styluses, as seen in Anoto pen, or employing IMU sensors as in the Digipen \citep{KIHT}. IMU-equipped styluses are versatile, allowing writing on tablets, paper, or boards. However, IMU signals only provide relative pen displacements, introducing potential noise.

Three categories of handwriting trajectory reconstruction emerge: i) reconstruction from offline handwriting images as \citep{chen2022complex, mohamedmoussa}; ii) reconstruction from pen-tip optical tracking systems as \citep{ott2022joint}; iii) reconstruction from IMU signals. Few works focus on IMU-based trajectory reconstruction. Some approaches \citep{bu2021handwriting, ott2022joint, wehbi2022surface}, use deep learning for online handwriting trajectory reconstruction as a step toward handwriting recognition. Earlier methods, like those in \citep{pan}, relied on traditional approaches like Hidden Markov models.

\begin{figure}[!ht]
\begin{center}
\includegraphics[width=\textwidth]{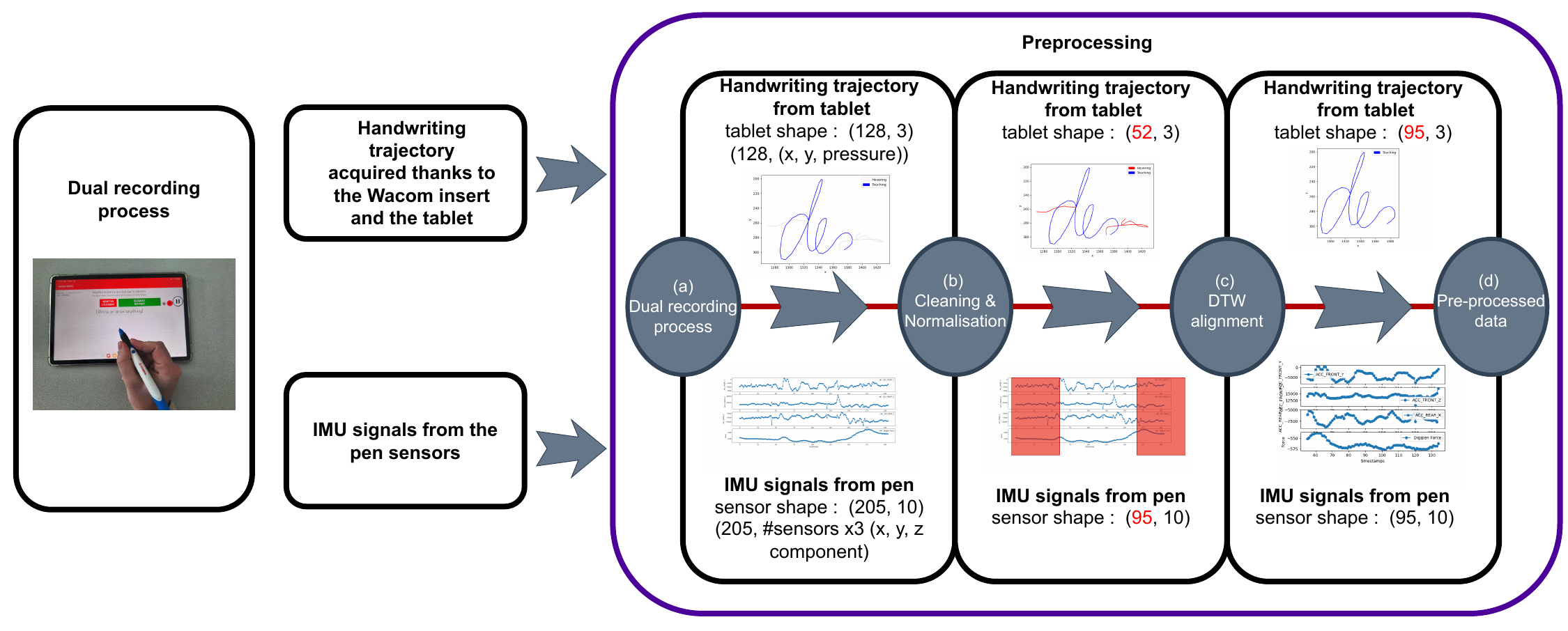}
\caption{The processing pipeline proposed in \cite{Swaileh2023}. (a) thanks to dual acquisition, they recover Digipen signals and the ground truth, (b) they remove start and end hovering, which are not data linked to handwriting, (c) they align ground truth and sensor signals using DTW, (d) preprocessed data used for training.}
\label{fig:sw}
\end{center}
\end{figure}

 \Florent{Recent advancements in handwriting trajectory reconstruction and IMU signal enhancement have highlighted several innovative approaches. \citep{bu2021handwriting} developed a handwriting assistant system that achieves millimeter-level accuracy by utilizing attachable inertial sensors, demonstrating the high precision achievable with IMU-based methods. Complementing this, \citep{pan} proposed a noise reduction technique specifically for IMU signals, which could significantly enhance the robustness of handwriting trajectory reconstruction models by addressing sensor noise challenges. In a related study, \citep{Shoes} introduced an unsupervised method for reconstructing pedestrian trajectories from IMU data, whose methodologies could be adapted to improve handling noisy signals during hovering strokes in handwriting applications. Furthermore, \citep{WangWavelet} presented a wavelet encoding network that enhances inertial signals, providing valuable insights for improving signal fidelity in handwriting datasets. \citep{Wang} also contributed with a GAN-based approach for inertial signal enhancement, which could inspire enhancements in reconstruction accuracy, especially for complex handwriting trajectories. \citep{chenaerial} proposed a method for extracting and denoising high-resolution vehicle trajectories from aerial videos, significantly improving the precision of trajectory data in intelligent transportation systems. Lastly, \citep{Chenship} focused on ship trajectory reconstruction from AIS sensor data, successfully integrating deep learning techniques with wavelet analysis for optimal signal denoising and demonstrating effective applications in trajectory reconstruction tasks. Together, these studies represent significant strides in the field, offering methodologies and insights that could greatly benefit handwriting trajectory reconstruction efforts.} \cite{MagHacker} explored magnetic signals, while \cite{wehbi2022surface} addressed the Stabilo Digipen, using linear interpolation to align pen and tablet signals as preprocessing. An  alignment method is necessary due to the requirement of matching sequence sizes for point-to-point loss calculation. They use a Convolutional Neural Network for the online handwritting trajectory reconstruction. 
 Recently \citep{Swaileh2023} improved on \cite{wehbi2022surface} by proposing a complete pipeline to achieve online handwritting trajectory reconstruction cf. Fig. \ref{fig:sw}.
 In particular, Dynamic Time Warping (DTW) is used to align the ground truth with the input signal, which has the advantage of preserving the dynamics of handwriting, unlike  linear interpolation as used in \cite{wehbi2022surface}.  
 
\begin{figure}[!ht]
\begin{center}
\includegraphics[scale=0.5]{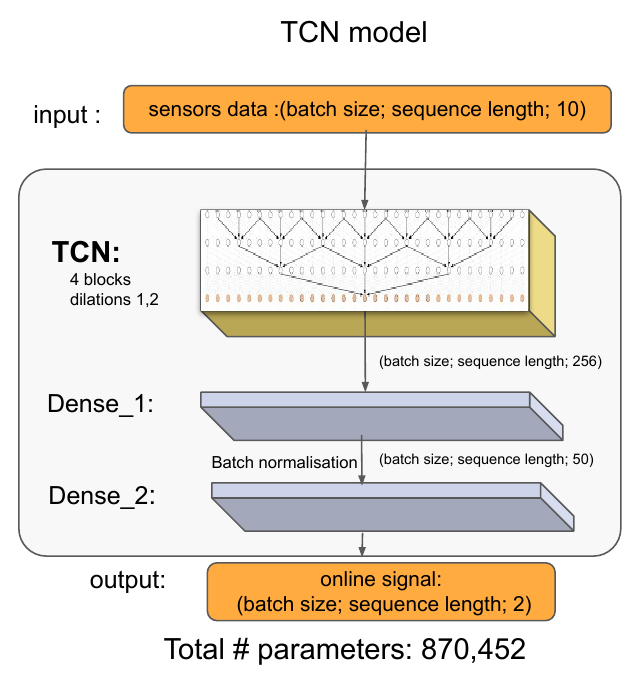}
\caption{TCN model proposed by \cite{Swaileh2023}}
\label{fig:TCN}
\end{center}
\end{figure}

They also propose the use of a Temporal Convolutional Networks (TCN) as a backbone network (Fig. \ref{fig:TCN}), which has the advantage of having larger receptive field than a classical CNN as used by \cite{wehbi2022surface} . 
 
 \cite{Swaileh2023} achieve very good results for the reconstruction of the touching parts of the handwriting, but the approach performs poorly for the reconstruction of the hovering parts.  Based on the model from \cite{Swaileh2023}, \cite{serdyuk} apply and compare different quantization techniques in order to enable the on-device inference of handwriting trajectory regression from inertial data.

\subsection{Using IMU Sensors for Handwriting Recognition}

While few works focus on handwriting reconstruction from IMU sensors, online handwritting recognition from IMU sensor data has been explored using pen-tip trajectory signals and IMU sensor signals. It should be noted that, although the two objectives may seem close, they are different in nature. The state-of-the-art shows that reconstructing from an IMU signal is a less complex task than reconstructing a precise trajectory. A benchmark study by \cite{ott2022benchmarking} compared neural network architectures for character, symbol, word, and equation recognition from IMU signals. CNN/BLSTM architecture proposed in \cite{ott2020onhw} showed the best results.
Some works address both online handwriting trajectory reconstruction and character recognition. \cite{wehbi2022surface} trained two neural networks sequentially for these tasks, assessing reconstruction quality based on character recognition rates. \cite{ott2022joint} proposed multitask learning using a CNN-LSTM network, demonstrating improved trajectory reconstruction and character classification. In the context of the UbiComp 2021 Challenge, \cite{wegmeth2021detecting} utilized a CNN/BLSTM to recognize mathematical expressions written with the Stabilo's Digipen, focusing on label boundary quality. Other studies minimize the link bandwidth between the Digipen and the remote device (e.g. tablet) \citep{kress2022hardware} or explore domain adaptation \cite{klass2022uncertainty, ott2022domain} and explainability \citep{azimi2022improving} in the context of online handwriting recognition from the Digipen. Recently, \cite{10318077} use low cost IMU from smartphone to  achieve characters recognition.  Their specificity is to couple inertial dynamic signals with trajectory morphology information coming from the handwriting image, then use a CNN for classification.

\section{Evaluation of Handwriting Reconstruction} \label{EHR}

Evaluation methods vary \Yann{from one study to another}, with some relying on qualitative assessments of reconstructed trajectories \citep{nguyen2021online}, while others use recognition performance as a proxy for evaluation  \citep{wehbi2022surface,huang2022agtgan}. Metrics like Root Mean Squared Error (RMSE), Dynamic Time Warping (DTW) and Fréchet distance are also commonly employed to assess reconstruction accuracy \citep{Swaileh2023, chen2022complex}. The diverse approaches underscore the complexity of evaluating handwriting trajectory reconstruction from digital stylus inputs.

\Florent{\Yann{In \cite{Swaileh2023}, the authors} stated that the Fréchet distance is a good metric to evaluate the trajectory reconstruction\Yann{. The Fréchet distance captures both local and global information accurately by taking into account temporal and spatial warping when aligning the predicted trajectory with the ground truth. It also} correlates well with qualitative assessment in practice.}

\Florent{Our experiments confirm this observation, as shown in Figure 4. \Yann{We have tested two experimental models and are observing the results for the different metrics. The} Fréchet distance aligns most closely with visual perception, particularly in reconstructing loops, such as those in letters like "f" and "e". This is also the case for "hinder", where the two reconstructions are \Yann{for us equivalent: in the first one, the loops and the characters are better made, but in the second one, the orientation is better. This results to close metric values, in contrast to the DTW values. Among various metrics, Fréchet distance uniquely captures the shape similarities. For this reason, we have chosen to only retain the Fréchet distance in our experiments.}
}

\begin{figure}[!t]
\begin{center}
\includegraphics[width=\textwidth]{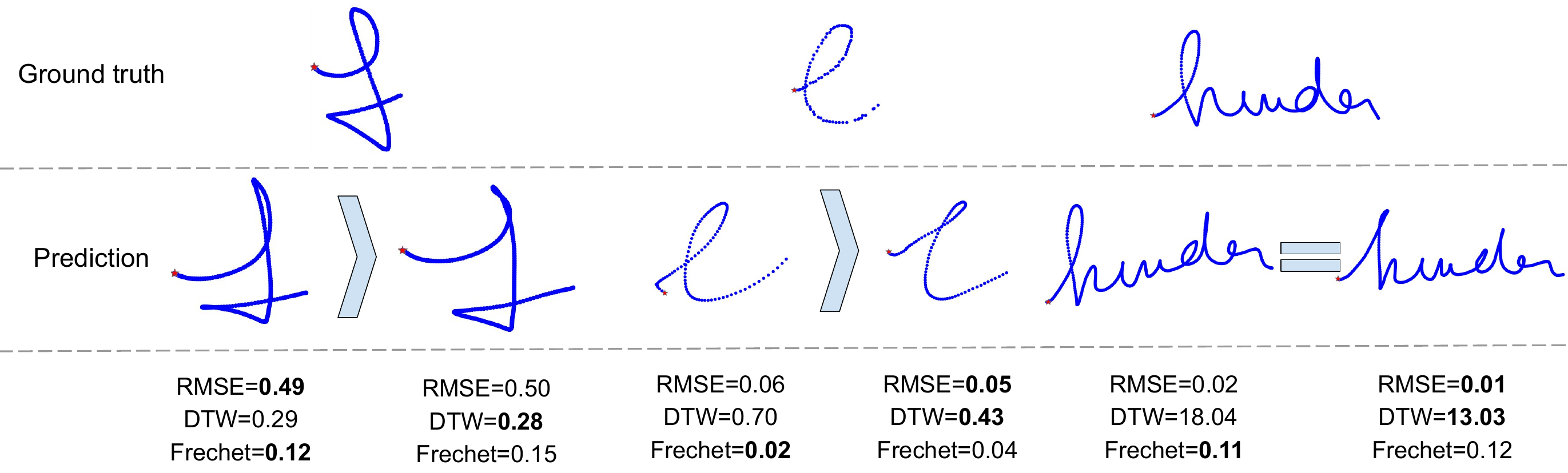}
\caption{Comparison between Euclidean distance, DTW and Fréchet distance \Yann{from predictions obtained using 2 different models. For the "f" and "e" characters, the predictions on the left seem better to us since the loops are better reproduced. For the word "hinder", the two predictions seem similar for us: the characters predicted on the left are better reproduced while the orientation is better on the right.}}
\label{fig:metrics}
\end{center}
\end{figure}

\Florent{
As a reminder, for two multivariate time series $A \in \mathbb{R}^{T_A \times z}$, and $B \in \mathbb{R}^{T_B \times z}$ of equal feature dimensionality $z$ and respective lengths $T_A$ and $T_B$, the Fréchet distance is defined by the following equation: 
\begin{equation}
F(A,B)= \min_{\delta} \max_{(i,j)\in \delta} d(a_i, b_j) 
\label{eq:frechet}
\end{equation}
}

%% file: Sections/Contributions.tex
\section{A new mixture-of-experts for better collaboration between 2D touching strokes and 3D hovering trajectories}
\label{Contributions}
%%\Yann{Suggestion de titre : A new mixture-of-experts for better collaboration between 2D touching strokes and 3D hovering trajectories}

\Yann{We propose an original mixture-of-experts dedicated to handwriting reconstruction. We first justify our proposal and then present our two expert models, including the training strategies.}

\subsection{\Yann{Motivation}}

\Florent{Reconstructing handwriting trajectories from inertial measurement unit (IMU) signals is a complex task that requires advanced algorithms to handle the multidimensional nature of the data. \Yann{Handwriting can be categorized into two phases: the act of writing (where the pencil makes contact with the surface) and the hovering phase (when the pencil transitions between two strokes).}} %Handwriting comprises two components: the touching part, which occurs in a 2D plane, and the hovering part, which exists in 3D space. 
\Yann{The touching part occurs in a two-dimensional plane as the act of writing is commonly on a flat surface while the hovering part is three-dimensional due to vertical variations of the pen. }
\Florent{The hovering component presents particular challenges due to the additional dimension, leading to greater variability in trajectories depending on the individual. Moreover, this phase can exhibit increased randomness because gestures are not constrained by graphomotor movements, further complicating the reconstruction process.}

\Florent{Inspired by the work of \cite{Swaileh2023}, which demonstrated success in \Yann{reconstructing} touching strokes, we \Yann{propose a new approach for dealing both with touching and hovering parts to reconstruct the entire trajectory of handwriting gestures.} %to handwriting trajectory reconstruction. 
Our methodology \Yann{is a mixture-of-experts that} focuses on breaking the reconstruction process \Yann{down} into two tasks, \Yann{the touching and hovering ones, resulting in} a more nuanced understanding of each phase of handwriting. %Handwriting can be categorized into two phases: the act of writing (where the pencil makes contact with the surface) and the hovering phase (when the pencil transitions between two strokes).
}

\begin{figure}[!t]
\centering
    \includegraphics[width=\textwidth]{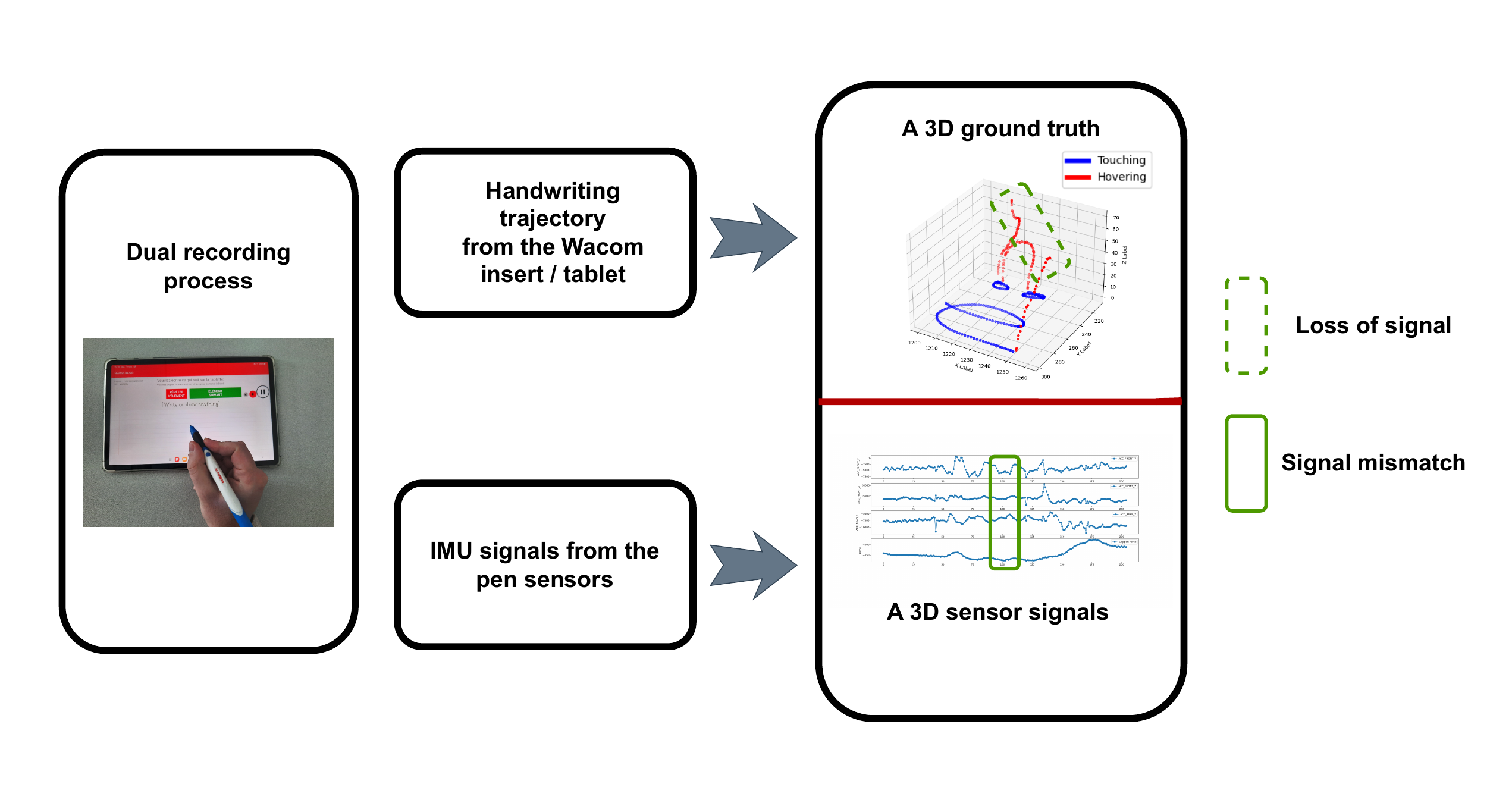}
    \caption{Dual recording process, with a Wacom insert, to enable the acquisition of ground truth.  Composed of writing (blue) and hovering parts (red) that are not always tracked (green). }
    \label{fig:rec_process}
\end{figure}

\Florent{The writing phase provides a two-dimensional ground truth based on the observable path of the pencil on the writing surface, making it amenable to direct measurement. In contrast, the hovering phase involves movement in three-dimensional space, during which the pencil must be positioned at the next stroke’s starting point. This phase is challenging because it lacks a clear ground truth if the pencil is lifted too high. As depicted in (Fig. \ref{fig:rec_process}), the absence of precise 3D labels during training limits our ability to accurately model this phase.}

\Florent{To address these challenges, our proposed MOE (Mixture-of-Experts) model effectively differentiates between the two phases of handwriting, allowing for targeted learning and improved trajectory reconstruction. By leveraging distinct approaches for the writing and hovering phases, our model aims to enhance accuracy while accommodating the variability inherent in human handwriting. \Yann{To acquire ground-truth trajectory data corresponding to the IMU signals, we employ the Digipen stylus equipped with a Wacom insert as illustrated in Figure~\ref{fig:rec_process}.} }

\begin{comment}

\begin{figure}[!t]
\centering
\begin{subfigure}[b]{0.28\textwidth}
    \includegraphics[width=\textwidth]{Images/GT/plot_figure_210.png}
    %\caption{"A" for ground truth with hovering}
\end{subfigure}%
\begin{subfigure}[b]{0.3\textwidth}
    \includegraphics[width=\textwidth]{Images/GT/plot_figure_597_bis.png}
    %\caption{an "ë" for ground truth with hovering, but part of it (green arrow) is not tracked}
\end{subfigure} \\
\begin{subfigure}[b]{0.3\textwidth}
    \includegraphics[width=\textwidth]{Images/GT/plot_figure_2327.png}
    %\caption{a "draw" for ground truth with hovering, but part of it (green arrow) is not tracked}
\end{subfigure}
\begin{subfigure}[b]{0.3\textwidth}
    \includegraphics[width=\textwidth]{Images/GT/plot_figure_1866.png}
    %\caption{an equation for ground truth with hovering, but part of it (green arrow) is not tracked}
\end{subfigure}
\caption{Example of data used as ground truth. Composed of writing (in blue on pictures) and hovering parts (red on pictures) that are not always tracked (green arrow).}
\label{fig:GT}
\end{figure}

\begin{figure}[!t]
\centering
    \includegraphics[width=\textwidth]{Images/recording process_v2.pdf}
    \caption{Dual recording process, with a Wacom insert, to enable the acquisition of ground truth. }
    \label{fig:rec_process}
\end{figure}
\end{comment}

%\Florent{To address these challenges, we employ the Digipen pen, equipped with a Wacom insert, to acquire ground-truth trajectory data that corresponds to the IMU signals (Fig. \ref{fig:rec_process}). Our proposed MOE (Mixture of Experts) model effectively differentiates between the two phases of handwriting, allowing for targeted learning and improved trajectory reconstruction. By leveraging distinct approaches for the writing and hovering phases, our model aims to enhance accuracy while accommodating the variability inherent in human handwriting.}\ 

\subsection{Our mixture-of-experts for handwriting reconstruction}

In this section, we present the overall mixture-of-experts we propose and discuss in details the design of the two experts. 

\subsubsection{Presentation of our mixture-of-experts}

We can formalize our problem as multitask learning, in that we have two linked tasks, the first being the prediction of the writing itself, the second the hovering movement between these different parts. These two tasks differ in their dynamics and nature (2-dimensional and 3-dimensional signals). 
%In practice, this means combining two specific neural networks.
\Yann{Here, we propose to have 2 expert models, each specialising in its own task. }
Inspired by \cite{Swaileh2023} where a TCN-based architecture is trained on touching strokes only producing degraded reconstructions on hovering parts, we suggest to use this network architecture as \Yann{a basis for our two expert models}. In the following, we will refer to this TCN-based model as the backbone model.

\begin{figure}[!ht]
\begin{center}
\includegraphics[scale=0.24]{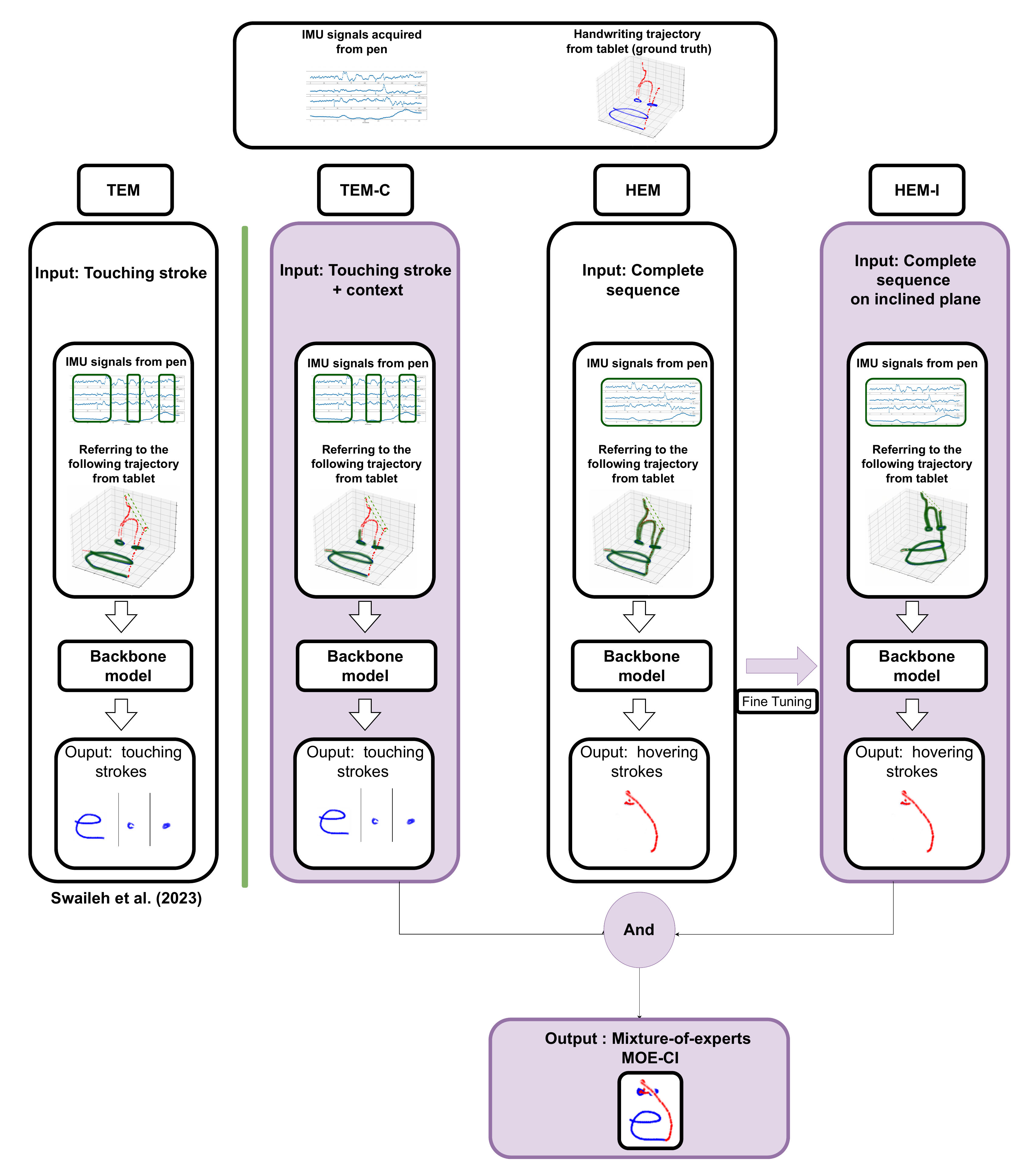}
\caption{{\footnotesize Our MOE-CI approach combines two models and their enhancements: TEM-C and HEM-I. TEM-C integrates hovering (in red) into the Touching Expert Model (TEM), providing better context and smoother transitions to touch. HEM-I fine-tunes the Hovering Expert Model (HEM) with 3D data for improved spatial understanding.}}
\label{fig:train}
\end{center}
\end{figure}

\Yann{We propose a new Mixture-Of-Experts (MOE) dedicated to handwriting reconstruction with two neural networks in parallel, one expert touching model and one hovering expert model. The mixture of models with the proposed training approach is illustrated in Figure~\ref{fig:train}.}

\Florent{The result is a global reconstruction of the handwriting. This approach functions as a mixture-of-experts model, with one expert focused on the handwriting during pen contact and the other handling the repositioning of the pen for the next stroke during hovering. The nature of the signals differs significantly between the constrained writing dictated by the graphomotor gesture and the repositioning trajectory during hovering. Unlike traditional MOE models, where switching between experts is predefined or data-driven, our model controls this transition dynamically through the pen's pressure sensor. This allows each expert to specialize more effectively, with the handwriting expert refining stroke precision and the hovering expert optimizing trajectory transitions.}

We call this approach MOE-CI (Fig \ref{fig:train}), which is the combination of the 2 following experts: 
\begin{itemize}
    \item the TEM-C for Touching Expert Model with hovering Context: corresponds to the backbone model trained with touching stoke with hovering portions preceding the touching strokes, it's an enhancement of the Touching Expert Model (TEM) which corresponds to the \cite{Swaileh2023} method; it is dedicated to touching prediction;
    \item the HEM-I for Hovering Expert Model with Inclined data fine tuning, corresponds to the backbone model trained with complete sequence named Hovering Expert Model (HEM), followed by a fine tuning to data acquire on inclined plan; it is dedicated to the hovering prediction;
\end{itemize}

\subsubsection{Network architecture}
We remain convinced that \Yann{a TCN-based network} is the right network architecture to handle these IMU data. In addition, it has a number of advantages, including the fact that it has fewer vanishing gradient problems than LSTMs, is faster to train, and remains a lightweight architecture that requires considerably less data in training than a Transformer. \Yann{This is particularly important in our use-case where the data is very specific and therefore relatively small in quantity. In addition, }one of our final goal is to embed the model in the pen, which requires a lightweight architecture. 

\Florent{As a reminder and for reproductibility purposes, the backbone model (Fig. \ref{fig:TCN}) is based on 4 blocks of a non-causal TCN followed by two dense layers, with a batch normalization layer applied between them. Each TCN block is composed of 2 convolutions with dilation 1 and 2 respectively, a kernel size of 3 and a dropout rate of 0.2.}
\Yann{Thus, our two expert models, i.e. the touching expert and the hovering expert, are based on this network architecture. } 
\Florent{The network is trained conventionally, with the MSE as the cost function and early stopping at 25 epochs. 10 recordings are used for the test set, with the training data accounting for 90\% of the remaining data and the rest for validation. }

\subsubsection{Models' inputs}
We therefore turned our attention on how to train our models, in particular with regard to the type of data given as input to the network. \Yann{Regarding the touching expert, the model is only trained on touching strokes, corresponding to 2D input signals. We believe that 3D signals from hovering parts may deteriorate the touching reconstruction due to the additional dimension that varies. Thus, the input signals are split into strokes according to the pen pressure and only the 2D signals from touching strokes are given in input of the touching expert model. }

Addressing the specificity of hovering strokes, we suggest to use the backbone model on complete sequences, because we believe that giving as much context as possible can be beneficial for hovering prediction. The reason for training our network on entire sequences, rather than isolating hovering strokes, comes from the complex dynamics of pen-to-tablet interactions. By having the full sequence, the model gains insights into the transition patterns between active stylus contact and hovering states. 

\Florent{Thus, TEM is trained on touching strokes, while HEM is trained on entire sequences. Additionally, we propose two variants: TEM-C and HEM-I, which differ in terms of input data. These variants are discussed in detail below.}

\subsection{TEM-C: Incorporating  temporal  context  that  reflects  physics  and  dynamics to enhance the touching expert model} \label{TEM-C}

\Florent{In the reconstruction of a trajectory based on Inertial Measurement Units, the temporal context plays a crucial role in accurately capturing the dynamic movement of an object. IMU are sensor systems that measure specific forces and angular rates to determine the acceleration and orientation of an object. Mathematically, the integration of these measurements over time helps to reconstruct object trajectory. 
The importance of temporal context in trajectory reconstruction process is the result from the necessity to accurately model continuous variations in acceleration and angular velocities, which reflect object dynamic behaviors including acceleration and directional shifts. By integrating temporal information, the object motion can be more accurately reconstructed, accounting for these dynamic changes.}

\begin{figure}[!ht]
\begin{center}
\includegraphics[width=\textwidth]{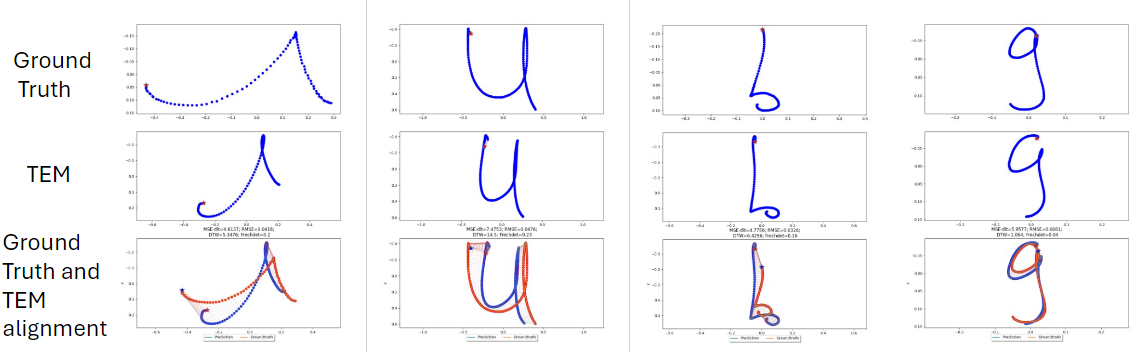}
\caption{On the first line the ground truth. On the second the TEM \cite{Swaileh2023} prediction. The observation is that the first points are less well reconstructed. On the last line the alignment between ground truth and the prediction.}
\label{fig:start}
\end{center}
\end{figure}

\Florent{This weakness has been confirmed in our experiments. The prediction was less accurate (Fig. \ref{fig:start}) for the initial points of a stroke due to a lack of dynamics.}

\begin{figure}[!ht]
\centering
\begin{subfigure}[b]{0.3\textwidth}
    \includegraphics[width=\textwidth]{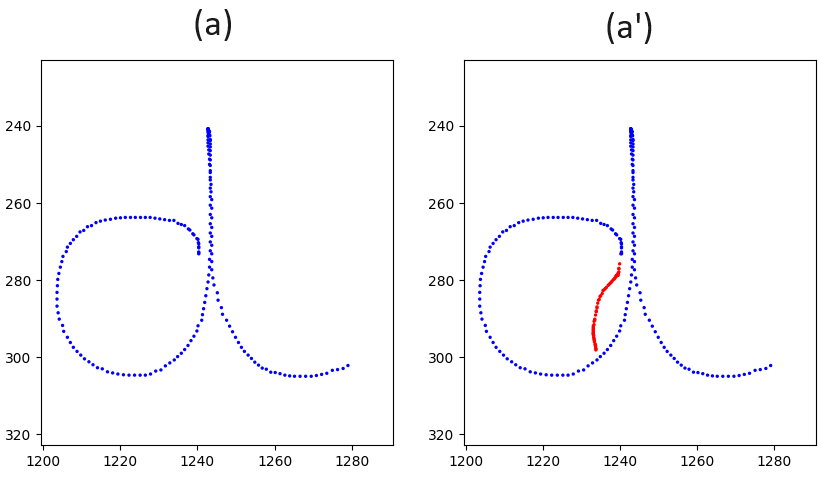}
\end{subfigure}
\begin{subfigure}[b]{0.3\textwidth}
    \includegraphics[width=\textwidth]{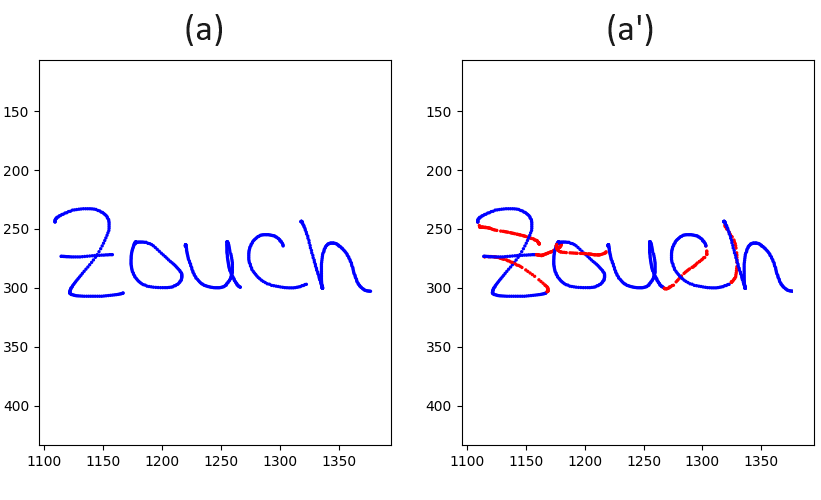}
\end{subfigure}
\begin{subfigure}[b]{0.3\textwidth}
    \includegraphics[width=\textwidth]{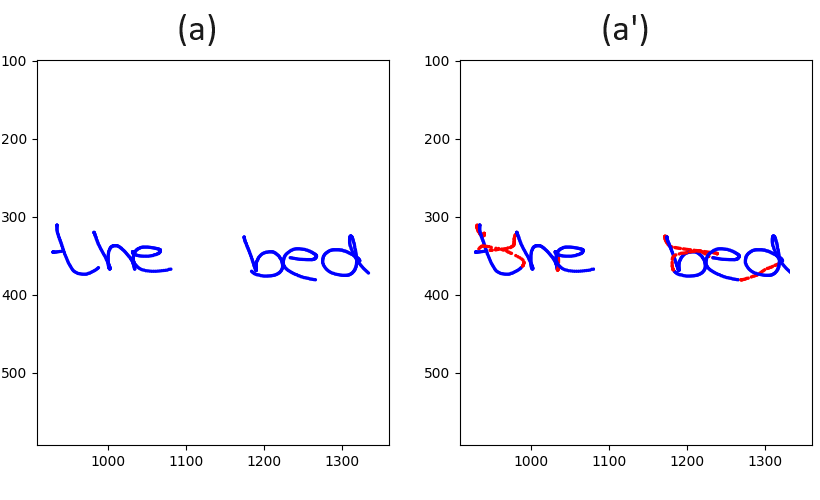}
\end{subfigure}
\caption{Visualizing the addition of hovering (in red) in (a') to ground truth in comparison to (a).}
\label{fig:addhov}
\end{figure}

\Florent{To take this physical aspect into account, dynamic context is given to input during the network training phase. 
For that, hovering movements preceding the touching strokes are added in the input sequences (in red on Fig. \ref{fig:addhov}). The size of this hovering movements corresponds to half of the receptive field that can be captured by the model on the left border of the touching strokes, so that the model sees no padding to predict the positions associated with the first touching values. }
\Florent{In addition, this will enable the network to see real signals instead of padding and thus to have a better generalization capability.}

\Florent{This effect on convolution padding can be translated mathematically as follows: let $x$ be the input of a TCN convolutional layer, $x$ is of dimension $T$ × $D^I$ where $T$ is the sequence length, and $D^I$ relates to the number of channels. 
Let $f$ be a filter (convolutional kernel) of size $T^f$ × $D^I$.}
\Florent{To preserve the input size in output, one considers a stride $s = 1$. \Yann{Traditionally, when preserving the input size, the input sequence is padded on the borders by adding zeros on the left and on the right, which corresponds to $\lfloor \frac{R^f}{2} \rfloor$ zeros on the left and $\lfloor \frac{R^f}{2} \rfloor$ on the right added to the input sequence,} where $R^f$ corresponds to the receptive field of the TCN. 
Here, we introduce hovering part instead of padding to prevent padding effects on the touching prediction on the left border.
We therefore have the following dilated convolutions:}
\begin{equation}
   y(i) =  \sum_{k=-R_p}^{R_p} \sum_{d=0}^{D^I} f(k+R_p, d)\: x(i+r \times k,d)
    \label{eq:conv}
\end{equation}
\Yann{where $R_p = \lfloor \frac{R^f}{2} \rfloor$ for a standard odd-sized kernel, $r$ is the dilation rate and $y(i)$ is the output at position $i$. }
\Yann{In a standard convolution, Equation~\ref{eq:conv} involves that $i + rk \geqslant 0$ ensuring valid indexing due to zero padding, whereas here, $i + rk$ is always valid due to the prior context from the pen's trajectory. This prior context is always available since there is always a downward trajectory of the pen before the first pressure point.  
}

\subsection{HEM-I: training on 3D labelled samples} \label{HEM-I}

\begin{figure}[!ht]
    \centering
    \begin{subfigure}{0.2\textwidth}
        \centering
        \includegraphics[width=\linewidth]{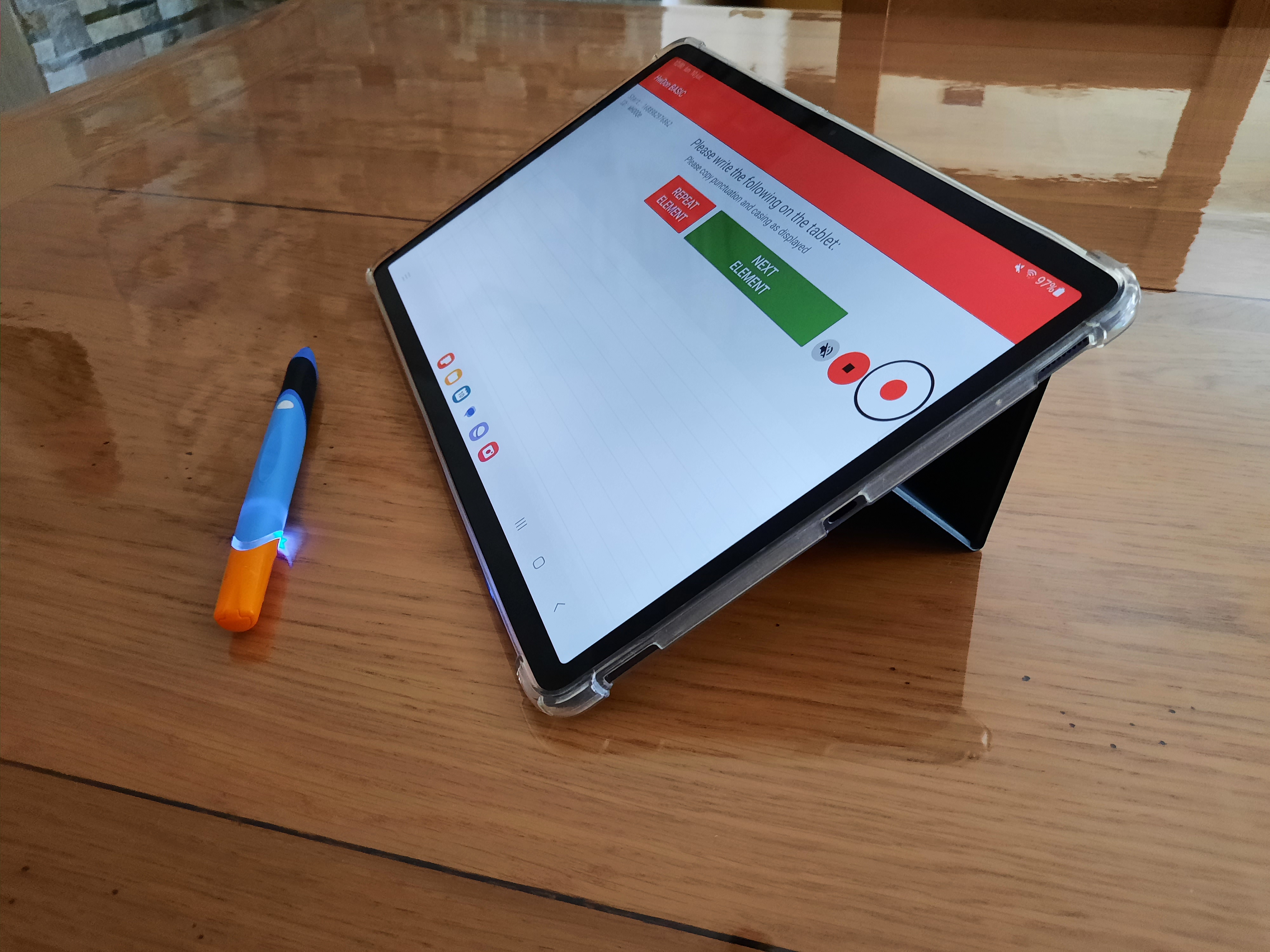} 
        \caption{Horizontal tablet inclined upward}
        \label{fig:sub1}
    \end{subfigure}
    \hfill 
    \begin{subfigure}{0.2\textwidth}
        \centering
        \includegraphics[width=\linewidth]{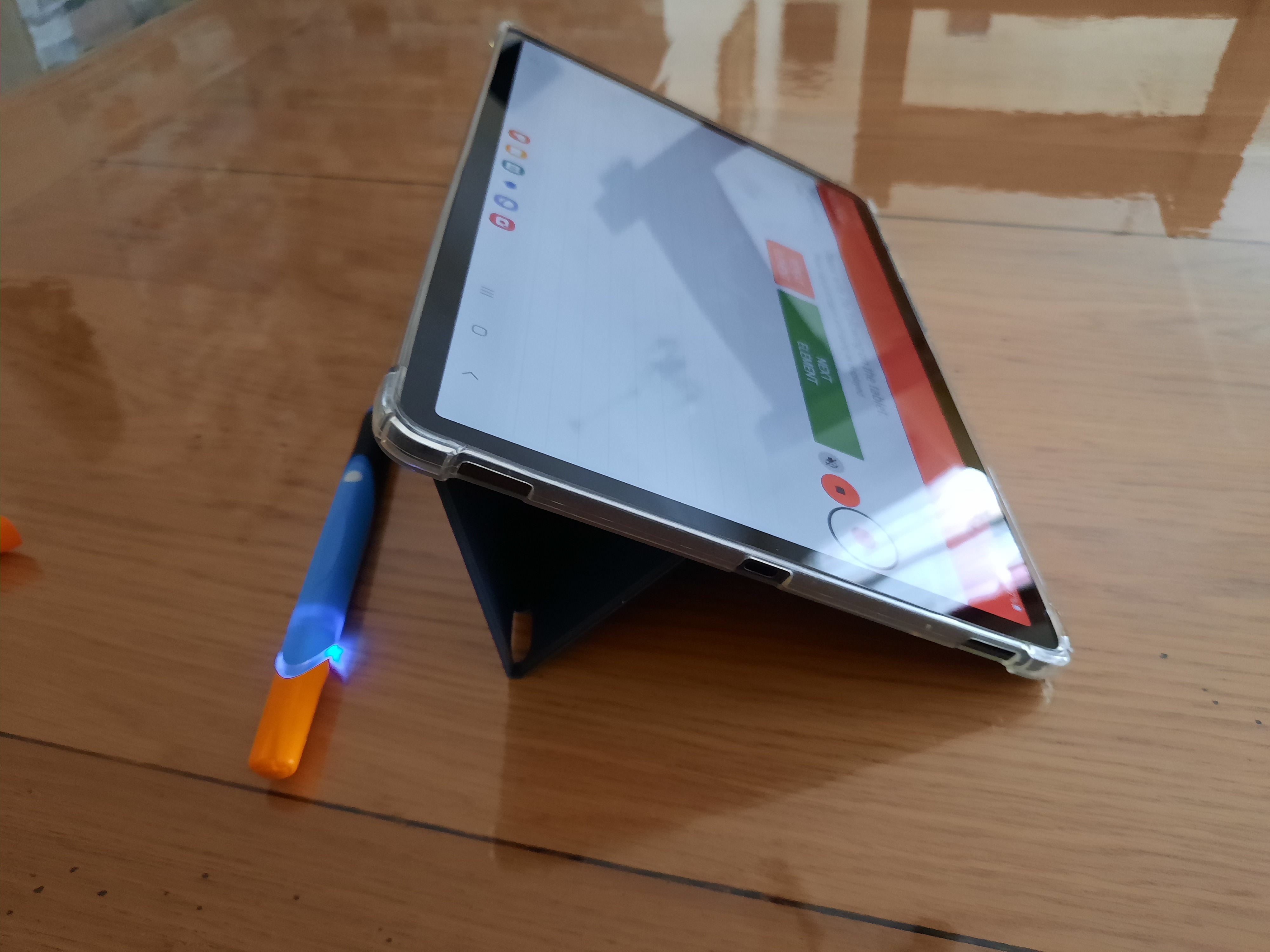}
        \caption{Horizontal tablet inclined downward}
        \label{fig:sub2}
    \end{subfigure}
    \hfill 
    \begin{subfigure}{0.2\textwidth}
        \centering
        \includegraphics[width=\linewidth]{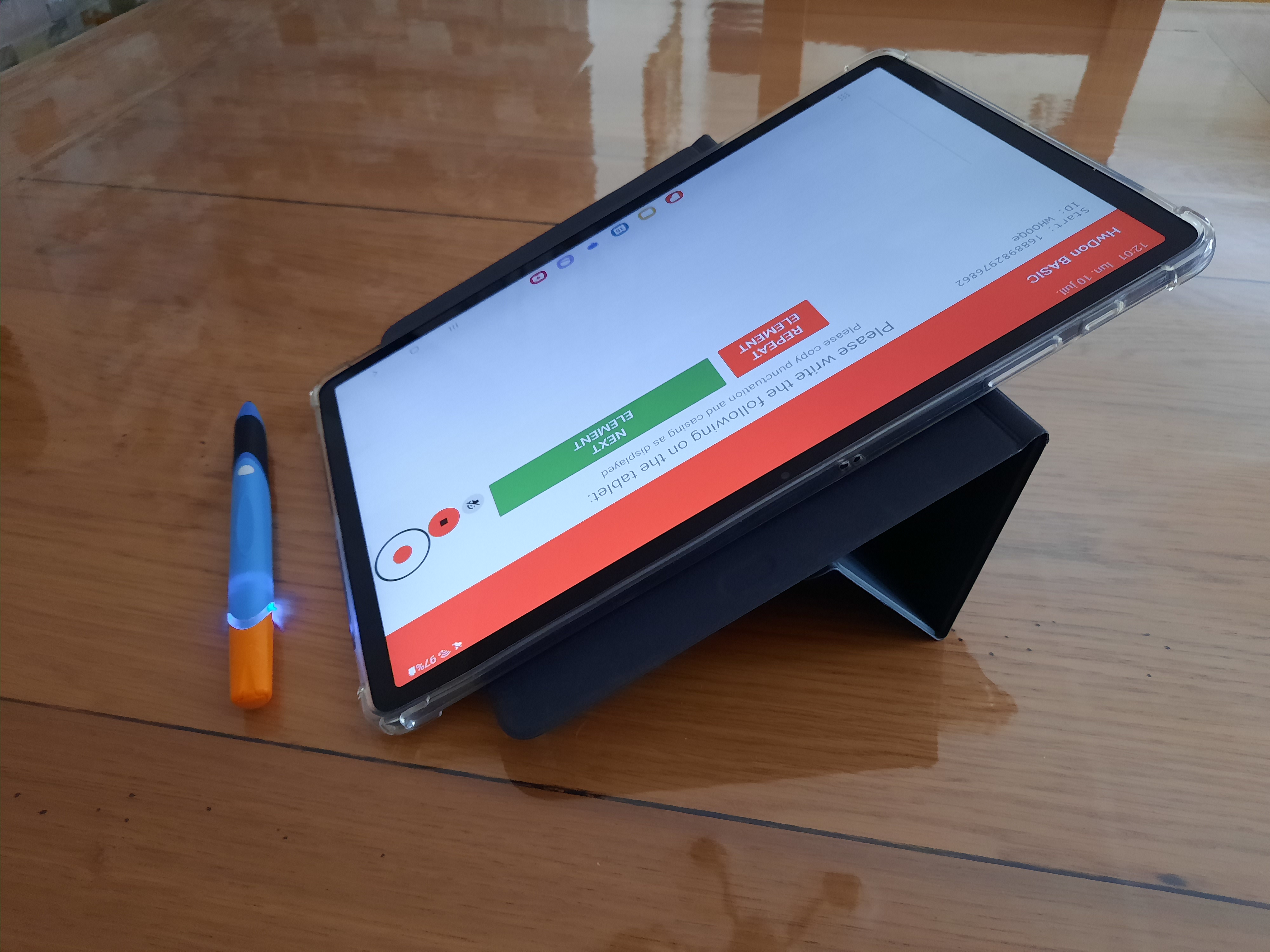}
        \caption{Vertical tablet inclined upward}
        \label{fig:sub3}
    \end{subfigure}
    \hfill 
    \begin{subfigure}{0.2\textwidth}
        \centering
        \includegraphics[width=\linewidth]{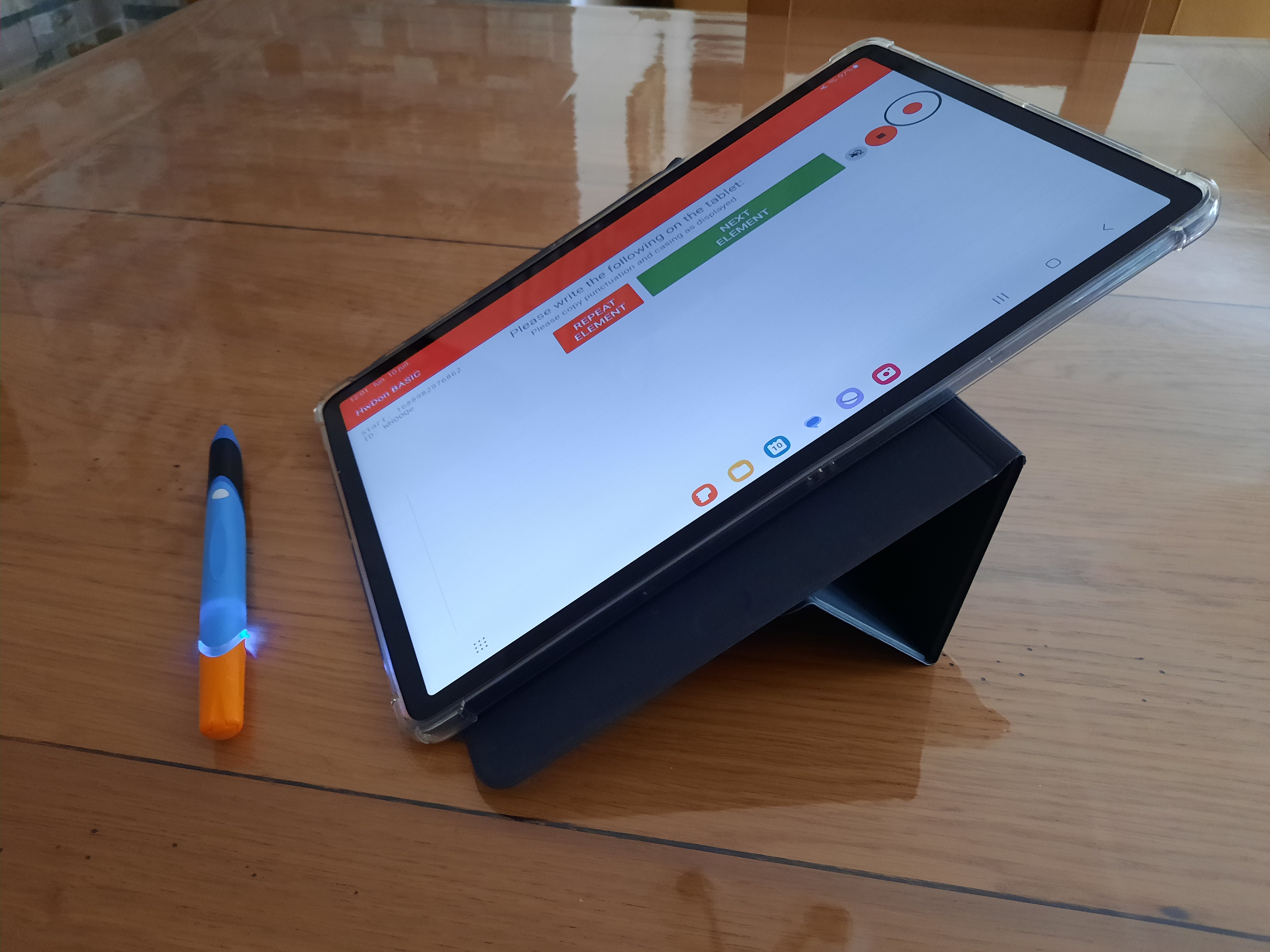}
        \caption{Vertical tablet inclined downward}
        \label{fig:sub4}
    \end{subfigure}
    \caption{Data acquisition protocol on inclined planes}
    \label{fig:set-up}
\end{figure}

\Florent{Handling the hovering phase in trajectory reconstruction presents unique challenges due to distinct dynamics compared to writing segments. While the touching parts share a common 2D plane, hovering movements introduces a third dimension, representing the height of the hovering movements, adding complexity to sensor data as well as the variability of unconstrained hovering trajectories. To address this variability, we rely on a dedicated network, called Hovering Expert Model (HEM). This model shares the same architecture as our TEM. It is pretrained on data containing both touching strokes and hovering phases and fine-tuned using data acquired on inclined planes to benefit from variations in the 3 dimensions at training time.
In this way, we expect that the fine-tuned network demonstrates enhanced adaptability to variations in hovering height, resulting in more robust predictions for hovering segments. 
Acquisition protocol to acquire inclined examples (Fig. \ref{fig:set-up}) includes several positions to introduce variability in the inclination of the writing surface. Four setups are considered, positioning the tablet horizontally with a 30-degree upward or downward inclination, and vertically with similar 30-degree inclinations upwards or downwards.} \\

%% file: Sections/Results.tex
\section{Results}
\label{Results}

We experiment our approach on two datasets. One is public to be used for research work comparisons, the other is private with more data. First, we report a comparative analysis of our novel mixture-of-experts (MOE) against the established approaches of  \cite{Swaileh2023} and \cite{wehbi2022surface}.  Then, we focus on the specific contribution of each expert model and their impact on overall reconstruction through an ablation study. We perform a comparative exploration of possible mixtures, in order to show the benefit of the proposed improvements for each expert into the collaborative mixture-of-experts. Finally, experiments are done on both the private and public databases, highlighting the impact of data quantity and diversity.

\subsection{Data sets description}

\begin{figure}[!ht]
\centering
    \includegraphics[width=0.7\textwidth]{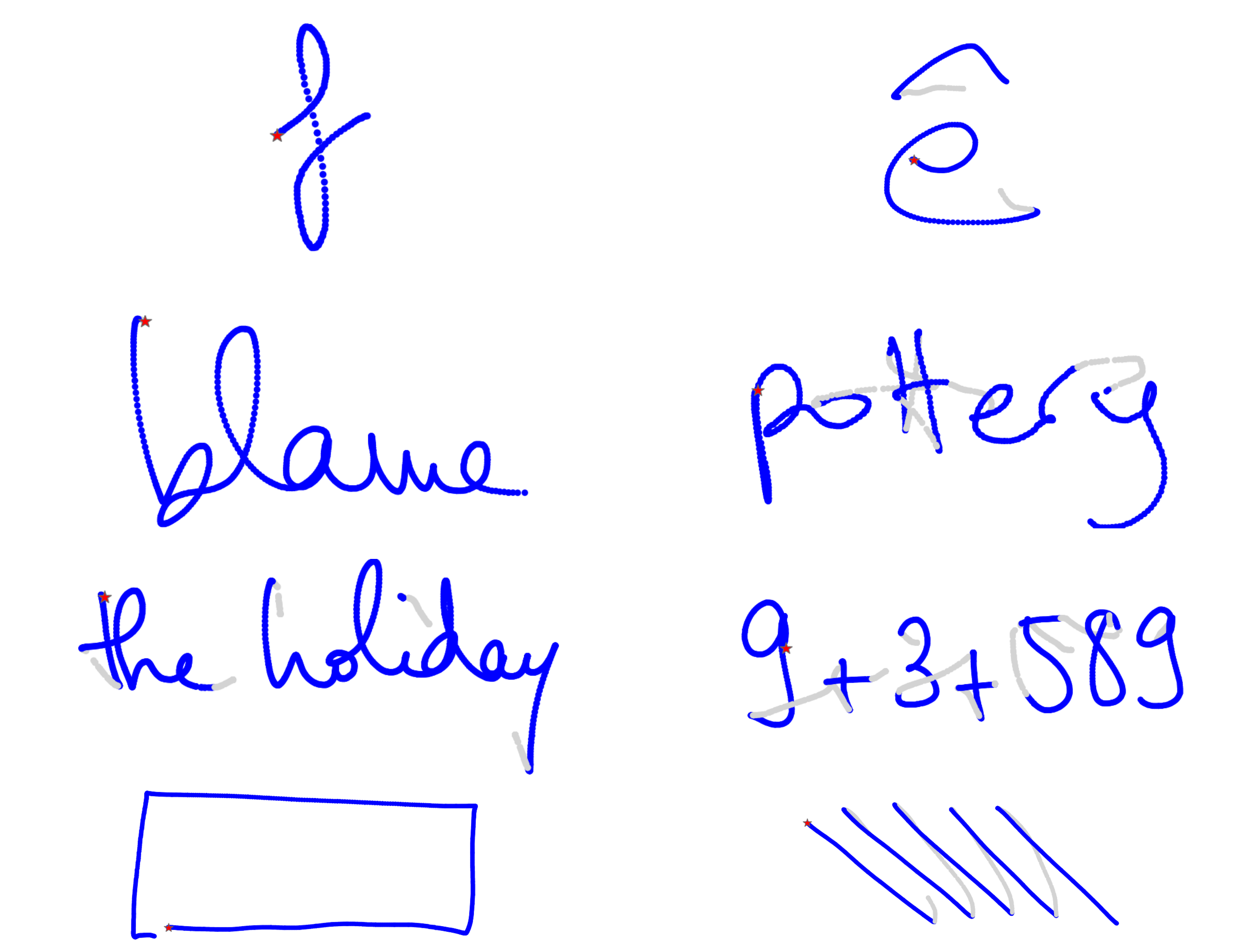}
    \caption{Some examples from the KIHT-Public dataset. Pen-up (hovering) strokes which are tracked by the tablet (i.e. under 7mm) are in gray and pen-down (touching) strokes are in blue.}
    \label{fig:data}
\end{figure}

In this work, we test our approach on two datasets, one public and one private. We can also note that these two datasets are extensions of those proposed by \cite{Swaileh2023}.
For this work we propose the KIHT-Public dataset, which contains 130 recordings. Each recording contains about thirty samples which are: characters, words, word groups, equations, and shapes (Fig. \ref{fig:data}). 
\Florent{The KIHT-Public dataset significantly surpasses the previous efforts by \cite{Swaileh2023} in terms of recording quantity and diversity, featuring 130 recordings as opposed to only 30. This substantial increase allows for a greater number of unique writers, introducing more variability in handwriting styles. This is crucial for developing robust recognition models. Additionally, the inclusion of inclined recordings introduces further variability, particularly in pencil heights during writing, which can affect the dynamics of the strokes. This feature not only enriches the dataset but also makes it suitable for handwriting reconstruction tasks.}

This dataset is publicly accessible\footnote{\url{https://www-shadoc.irisa.fr/irisa-kiht-s-and-kiht-public-datasets/}} for research purposes.
We have also evaluated our work on a private dataset (denoted KIHT-Private), which is an extension of the KIHT-Public one with 300 additional recordings. Details on datasets are available in Table \ref{tab:datasets}. Note that the test recordings are the same for these two datasets to allow fair comparisons.

\begin{comment}
\begin{table}[!ht]
\centering
\caption{Important stats of the considered datasets}
\label{tab:datasets}
\footnotesize
\begin{tabular}{|l|l|l|p{2.3cm}|p{2.5cm}|}
\hline
Datasets & Sets & \# Writers & \# Recordings & \# Samples \\ \hline
KIHT-Public               & Training      & 36                  & 90                   & 2761              \\ \cline{2-5} 
& Inclined data          & 7                  & 30                     & 1368                  \\ \cline{2-5} 
                   & Test          & 9                  & 9                     & 266                  \\ \cline{2-5} 
                   & Total         & 46                  & 129                     & 4129                  \\ \hline
KIHT-Private            & Training      & 66                 & 371                   & 11811             \\ \cline{2-5} 
& Inclined data          & 12                  & 49                     & 2234                  \\ \cline{2-5} 
                   & Test          & 9                  & 9                     & 266                  \\ \cline{2-5} 
                   & Total         & 76                  & 429                     & 14045                  \\ \hline
\end{tabular}
\end{table}
\end{comment}

\begin{table}[ht]
\centering
\caption{Important stats of the considered datasets}
\label{tab:datasets}
\footnotesize
\begin{tabular}{|l|l|l|l|l|}
\hline
Datasets       & Sets         & \# Writers & \# Recordings & \# Samples \\ \hline
KIHT-Public    & Training     & 36         & 90            & 2761       \\ \cline{2-5} 
               & Inclined data & 7          & 30            & 1368       \\ \hline
KIHT-Private   & Training     & 66         & 371           & 11811      \\ \cline{2-5} 
               & Inclined data & 12         & 49            & 2234       \\ \hline
\multicolumn{2}{|l|}{Common test set} & 9   & 9             & 266        \\ \hline
\end{tabular}
\end{table}

\subsection{A new evaluation protocol}
\begin{figure}[!ht]
\begin{center}
\includegraphics[width=\textwidth]{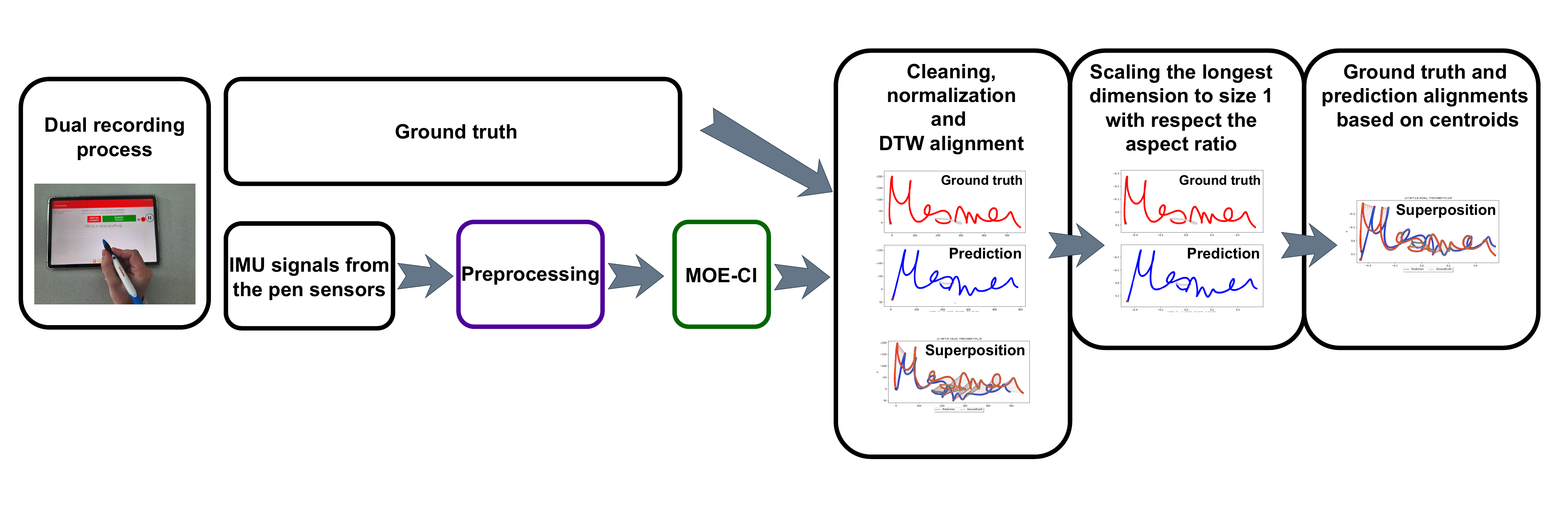}
\caption{Our evaluation pipeline, composed of four steps, dual acquisition of Digipen signals and ground truth, alignment with DTW to find identical sequence sizes, scaling and centering on centroids.}
\label{fig:eval}
\end{center}
\end{figure}

\cite{Swaileh2023} highlighted that the Fréchet distance serves as an effective metric for evaluating trajectory reconstruction, which is why we have incorporated this metric into our evaluation. Given our focus on the shape rather than the size of the reconstruction, we propose two additional steps before calculating the Fréchet distance, as depicted in Figure (Fig. \ref{fig:eval}).
The first step is to find  the longest dimension of the ground truth bounding box (resp. the reconstruction), and set its size to 1. The aim is not to give too much weight in the evaluation to the size of reconstruction, but to the overall quality of reconstructions. The second step consists in centering the centroids of the prediction and ground truth bounding boxes. Another difference with \cite{Swaileh2023} is that the Fréchet distance is not normalized by the sequence length in our evaluation protocol, so we can have a quantitative analysis of the impact of sequence length on reconstruction quality. 

We also propose a dual-level evaluation. The first one is named label level, here we're evaluating all the touching strokes of a label. This process allows us to assess the overall quality of the reconstruction, both writing parts and repositioning errors.
The second one is named stroke level and it refers to individual touching strokes only. It quantifies the quality of the handwriting reconstruction. 

\subsection{Comparison of our mixture-of-experts (MOE-CI) with state-of-the-art methods}

\begin{figure}[!ht]
\centering
    \includegraphics[width=0.8\textwidth]{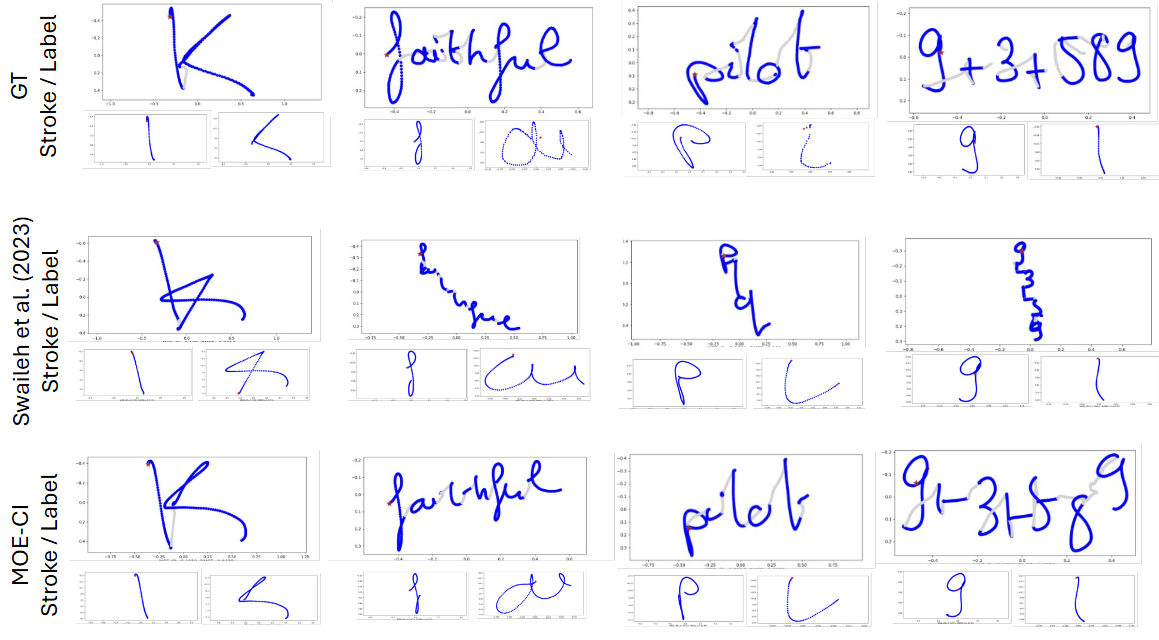}
    \caption{Comparison of our approach to that of \cite{Swaileh2023}, on the first line the ground truth at stroke (down) and label (up) level, on the second the reconstruction following the method of \cite{Swaileh2023}, on the last line our mixture-of-experts (MOE-CI).}
    \label{fig:eval_mix}
\end{figure}

%Using the evaluation protocol and datasets that have been presented, 
\Yann{We compared our MOE-CI approach to the work of \cite{Swaileh2023} and \cite{wehbi2022surface}, the two references in the field.} As a reminder, they proposed an approach based on a TCN model learned on touching strokes only for \cite{Swaileh2023} and a CNN model with a linear interpolation between sensor signals and ground truth for \cite{wehbi2022surface}. \Yann{We experiments these three approaches on the KIHT-Private dataset.} These results are presented qualitatively  (Fig. \ref{fig:eval_mix}) and quantitatively (Table. \ref{tab:eval_mix}). 
The Fréchet distances at label and strokes levels are significantly \Yann{lower (and better) with our mixture-of-experts}. We can see that on the touching parts, our approach reaches results close to those of \cite{Swaileh2023}. One recalls that \cite{Swaileh2023} has a model dedicated to touching strokes. 
\Florent{A model specifically designed for hovering enhances the accuracy of predictions, as illustrated in Figure \ref{fig:eval_mix}, where the repositioning between strokes is much more accurate than for the baseline method \cite{Swaileh2023}. Additionally, incorporating contextual information into the touching model significantly improves the prediction of touching parts. This is evident at the stroke level of “faithful,” where the loop in “ai” is better formed using the MOE-Ci method than with \cite{Swaileh2023} approach. Overall, the MOE-CI allows each model to be tailored to the two different signal natures, leading to improved understanding and reconstruction.}

\begin{table}[!ht]
\centering
\caption{Fréchet distance computed for our mixture-of-experts and state-of-the-art methods train on the KIHT-Private dataset, and evaluate on the test set. The line "\# wins" counts the number of labels where the model obtains the best scores against the others.}
\label{tab:eval_mix}
\footnotesize
\begin{tabular}{|l|c|c||c|}
\hline
Evaluation & \cite{wehbi2022surface} & \cite{Swaileh2023}  & Our: MOE-CI \\ 
\hline
Label level &  0.571           & 0.437                                    & \textbf{0.312}                        \\ \hline
Stroke level  &  0.120        & 0.097                               & \textbf{0.091}                        \\ \hline
\# Wins & 24        & 54                               & \textbf{188}                        \\ \hline
\end{tabular}
\end{table}

\subsection{Ablation study}
In this section, we focus on the ablation study to evaluate the impact of different contributions on the overall performance of our mixture-of-experts. 
We first explore the impact of the temporal context integration on the touching expert (TEM vs TEM-C) (cf. \ref{TEM}), and then the impact of the extra dimension on the hovering expert (HEM vs HEM-I) (cf. \ref{HEM}). This ablation study is performed on the KIHT-Private dataset.

\subsubsection{Temporal context integration on the touching expert (TEM-C)} \label{TEM}

\begin{figure}[!ht]
\centering
    \includegraphics[width=0.8\textwidth]{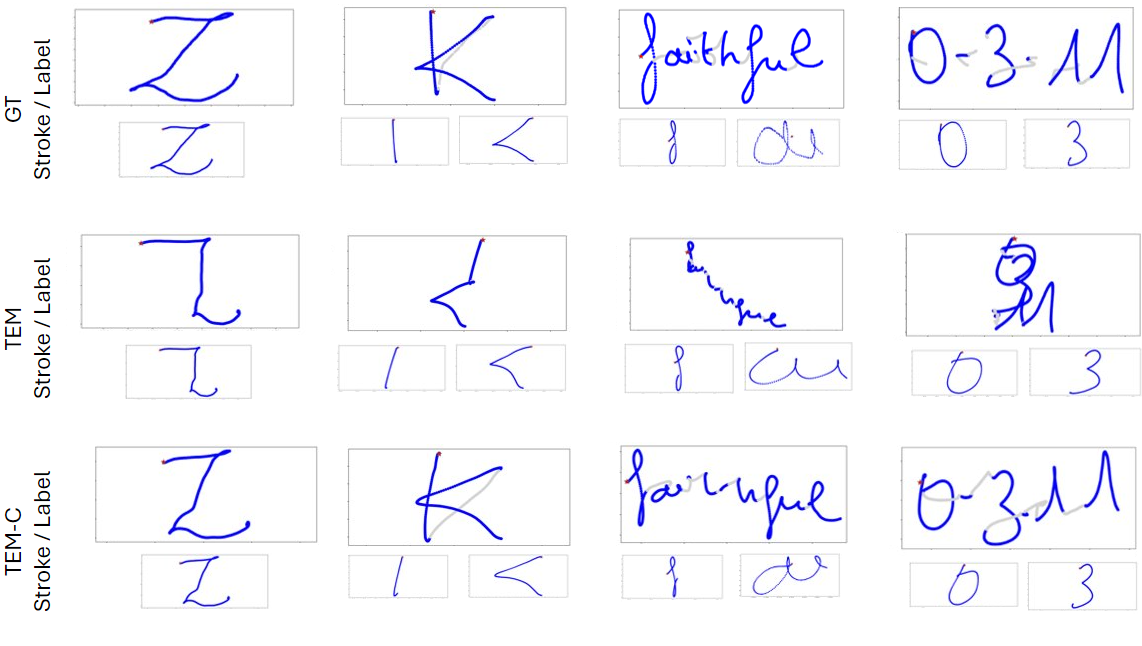}
    \caption{Comparison between the touching expert model (TEM) and its enhancement (TEM-C), on the first line the ground truth at stroke (down) and label (up) level, on the second line the expert model dedicated to touching strokes, on the last line the TEM improved by adding temporal context (TEM-C).}
    \label{fig:eval_addhov}
\end{figure}

\Florent{As presented in section \ref{TEM-C}, we suggest to integrate temporal context in input to the Touching Expert Model.
Results are shown at stroke level both quantitatively in Tab. \ref{tab:eval_addhov} and qualitatively in Fig. \ref{fig:eval_addhov}.
Adding temporal context significantly improves the model's ability to capture dynamic movement. By adding an earlier hovering portion to the touching strokes inputs, the model can now account for past dynamics, providing a more comprehensive understanding of the trajectory. This can be seen in Figure  \ref{fig:eval_addhov}, where the initial loop of the "a" in "faithful" is more accurately rendered, and the "l" is more precisely shaped. Similarly, the overall shape of characters such as the 'Z' and the '3' is better reconstructed.
It also appears to enhance stroke reconstruction and improve repositioning, as the model can see during training sections of 3D signals.}

\begin{table}[!ht]
\centering
\caption{TEM and TEM-C comparison at label and stroke levels using the Fréchet distance. The number of examples for which a model is better than the other is also computed.}
\label{tab:eval_addhov}
\footnotesize
\begin{tabular}{|l|c|c|c|}
\hline
Evaluation & TEM & TEM-C \\ 
\hline
Label level           & 0.437                 & \textbf{0.334}                        \\ \hline
Stroke level          & 0.097            &   \textbf{0.091}                        \\ \hline
\# Wins        & 69                           & \textbf{197}                        \\ \hline
\end{tabular}
\end{table}

\subsubsection{Fine tuning on extra dimension for the hovering expert (HEM-I)} \label{HEM}

\begin{figure}[!ht]
\centering
    \includegraphics[width=0.8\textwidth]{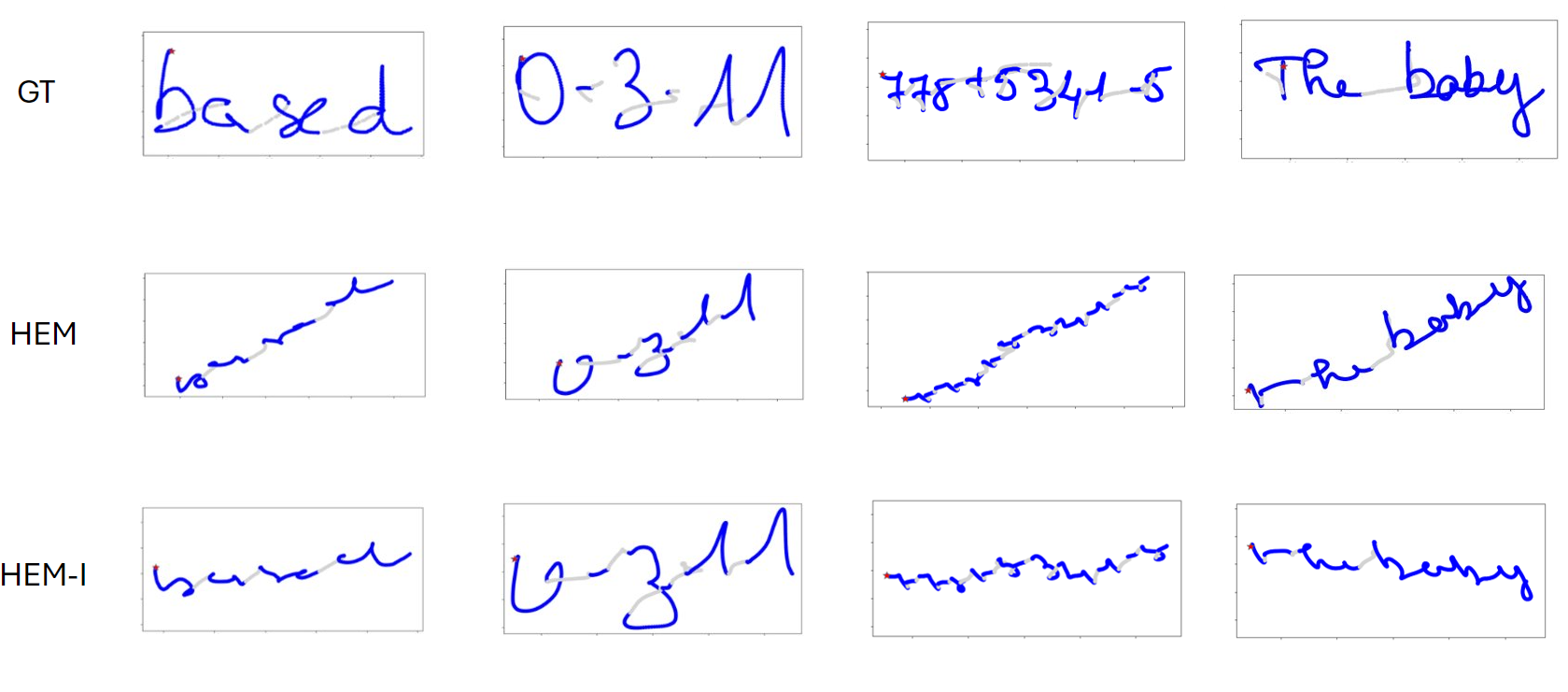}
    \caption{Comparison between the hovering expert model and with a fine-tuning on inclined data, on the first line the ground truth, on the second line the expert model dedicated to hovering (HEM), on the last line the fine-tuned model on inclined data (HEM-I).}
    \label{fig:eval_fine}
\end{figure}

We now evaluate the training of the hovering expert on inclined data, including the extra dimension (discussed in section \ref{HEM-I}). Fine-tuning the HEM model on data acquired from inclined planes has yielded remarkable improvements in prediction accuracy. 
It can be seen both at the stroke and label level, quantitatively in Table. \ref{tab:eval_fine} and qualitatively in  Fig. \ref{fig:eval_fine}.
Our expert model seems to tackle the inherent variability in sensors during hovering parts, that is due to different dynamics between writing and hovering strokes and from having the height dimension varying in the data.
\Florent{Our expert model effectively addresses the inherent variability in sensor data during hovering phases, which arises from the different dynamics between writing and hovering strokes, as well as fluctuations in the height dimension. By training on data from inclined planes to simulate various pencil heights, we enable the model to encounter a broader range of height variability, improving its ability to predict hovering segments. The fine-tuned model (HEM-I) exhibits adaptability to changes in pencil hovering height, resulting in a more robust framework for predicting these segments, as evidenced by the Fréchet distance measurements at the label level. This results in qualitatively better hovering prediction on all examples. Furthermore, improvements at the stroke level can be attributed to a more precise orientation of characters in their reconstructions, stemming from an enhanced understanding of spatial relationships.} 

\begin{table}[!ht]
\centering
\caption{HEM and HEM-I comparison at label and stroke levels using the Fréchet distance. The number of examples for which a model is better than the other is also computed.}
\label{tab:eval_fine}
\footnotesize
\begin{tabular}{|l|c|c|c|}
\hline
Evaluation & HEM & HEM-I \\ 
\hline
Label level           & 0.413                 & \textbf{0.350}                        \\ \hline
Stroke level          & 0.129            &   \textbf{0.114}                        \\ \hline
\# Wins        & 77                           & \textbf{189}                        \\ \hline
\end{tabular}
\end{table}

\subsubsection{Comparison of model combinations into a mixture-of-experts}
\begin{figure}[!ht]
\centering
    \includegraphics[width=0.8\textwidth]{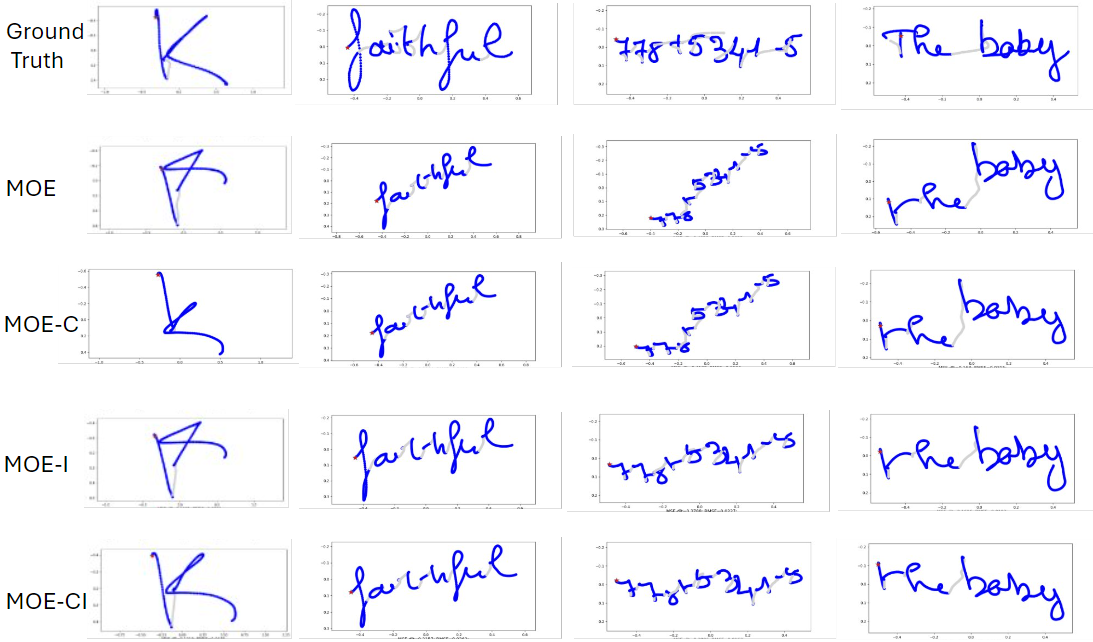}
    \caption{Comparison between the different combinations of mixture-of-experts, on the first line the ground truth, then from the top to bottom: MOE, MOE-C, MOE-I and MOE-CI. }
    \label{fig:eval_mixture}
\end{figure}

Now that each expert has been established, we evaluate the performance of each possible mixture combination (Fig. \ref{fig:eval_mixture} and Table \ref{tab:eval_mixture}). 
We introduce some notation for the mixture-of-experts (MOE) :
\begin{itemize} \label{mix}
    \item The combination touching expert model (TEM) \& hovering expert model (HEM) will be noted: MOE;
    \item The combination touching expert model (TEM) \& Hovering expert model fine tuned on inclined data (HEM-I) will be noted: MOE-I;
    \item The combination touching expert model with temporal context (TEM-C) \& hovering expert model (HEM) will be noted: MOE-C;
    \item As a reminder, the combination touching expert model with temporal context (TEM-C) \& Hovering expert model fine tuned on inclined data (HEM-I) will be noted: MOE-CI.
\end{itemize}

This evaluation shows us the benefits of different contributions within a mixture of models. The addition of context enables more accurate reconstruction of touching strokes, especially at their extremities. \Florent{This can be seen visually in the reconstruction of the second stroke of the ‘k’ or ‘a’ in ‘faithfull’. } Note that differences in performance may occur at stroke level for the same expert as the bounding box normalizations are applied at label level. Fine-tuning on 3D data enables better repositioning of hoverings. \Florent{This is particularly noticeable on long sequences with a lot of hovering, such as the equation where the hovering is much better reconstructed. But also on long hovering sequences, for example the hovering between ‘the’ and ‘baby’, which is much better reconstructed.} In addition, the combination MOE-CI, which is the combination of the 2 improved experts, is actually the method that shows the best results, as expected. \Florent{We can see that the reconstructions are better for two reasons: better prediction of hovering and an improvement in stroke.} The MOE-CI is more often better than the others, which attests to the benefits of our contributions. Nevertheless, the MOE has correct performance due to good performances on short mono stroke examples, which are the easiest part to reconstruct. 

\begin{table}[!ht]
\centering
\caption{Evaluation of possible mixture of expert on the KIHT-Private dataset. The Fréchet distance is computed to evaluate the models at label and stroke level. Note here the sum of the  \# Wins line is greater than the number of test labels, since there are common parts in the MOE, so in the event of a tie, the score for the 2 networks is counted.}
\label{tab:eval_mixture}
\footnotesize
\begin{tabular}{|l|c|c|c|c|}
\hline
Evaluation &  MOE & MOE-I &  MOE-C & MOE-CI  \\ \hline
Label level & 0.344 & 0.321 & 0.333& \textbf{0.312} \\ \hline
Stroke level & 0.096 & 0.091 & 0.096 & \textbf{0.091} \\ \hline
\# Wins & 93 & 86 & 85 & \textbf{123} \\ \hline
\end{tabular}
\end{table}

\subsection{Evaluation on the public dataset}

We have released the KIHT-public dataset that will serve as a benchmark for futue research works. 
We evaluate the performance of our mixture-of-experts (MOE-CI) and compare it both to the different expert combinations (MOE, MOE-C, MOE-I) and to the state-of-the art approaches \citep{wehbi2022surface,Swaileh2023}. 
The results (Table \ref{tab:eval_mixture_public}) are similar to those obtained on the private dataset. Indeed, the integration of an expert trained at the label level significantly improves performance at the label level compared to the state-of-the-art, and thus improves the processing of repositioning of the reconstructed handwriting after hovering. The proposals associated with each expert improve the robustness of mixture-of-experts. Whether it's incorporating temporal context that reflects physics and dynamics to enhance the touching expert model, or refining the hovering expert model with 3D labeled samples for improved hovering predictions.
These consistent results shows that the public dataset is relevant to be use as benchmark.

\begin{table}[!ht]
\centering
\caption{Fréchet distance computed for our mixture-of-experts and state-of-the-art methods train on the KIHT-Public dataset, and evaluate on the test set. The line "\# wins" counts the number of labels where the model obtains the best scores against the others.}
\label{tab:eval_mixture_public}
\footnotesize
%\begin{tabular}{|l|p{1.9cm}|p{1.9cm}||c|c|c|c|}
\begin{tabular}{|l|c|c||c|c|c|c|}
\hline
Evaluation & \cite{wehbi2022surface} & \cite{Swaileh2023}  &  MOE & MOE-I &  MOE-C & MOE-CI \\ \hline
Label level & 0.583& 0.464& 0.336 & 0.331 & 0.321 & \textbf{0.320} \\ \hline
Stroke level & 0.127 & 0.101 & 0.101 & 0.100 & 0.097 & \textbf{0.096} \\ \hline
\# Wins & 16 & 29  & 68 & 65 & 56 & \textbf{93} \\ \hline
\end{tabular}
\end{table}

%% file: Sections/Limitations.tex
\section{Limitations}
\label{Conclusion}

\Yann{Dependence on the public dataset may be problematic. Although the dataset has been significantly expanded compared to previous state-of-the-art works, it still cannot cover the spectrum of use cases, particularly in terms of unique handwriting styles and variability in input data. This could potentially limit the model's generalization capabilities across various real-world scenarios.}

\Florent{In addition, while the model demonstrates good ability to generalize across different users, it exhibits significant limitations in handling handwriting from children. \Yann{Indeed, children have less graphomotor skills, which produces input data that is very different from that of adults, on which the models were trained. In the same way, the model has been trained on tablet writing} and the inherent noise associated with paper-based writing significantly hinders its applicability in its current form.} 

%\Florent{The reliance on a public dataset presents another challenge, as the dataset's focus on specific handwriting styles may not adequately represent a broader spectrum of use cases, potentially limiting the model's generalizability to varied handwriting styles encountered in real-world scenarios.} 

\Florent{Furthermore, the architecture comprises two dedicated expert models, which \Yann{%, while maintaining computational efficiency,
requires more memory, limiting the use in an embedded version within the stylus.} This presents potential computational challenges in real-time applications, particularly when aiming for deployment in memory-constrained environments \Yann{while maintaining computational efficiency}.}

%% file: Sections/Conclusion.tex
\section{Conclusion \& perspectives}
\label{Conclusion}

This paper introduces a novel method for reconstructing handwriting trajectories from kinematic sensor signals embedded in a digital pen, called Digipen designed by STABILO. 
We present a mixture-of-experts where each model is task-specific. The first expert model predicts touching strokes and processes 2-dimensional signals. The second expert model predicts trajectory repositioning between two touching strokes and processes 3-dimensional inputs due to pen-up movements. 
We demonstrated the relevance of our proposed improvements for each expert model. We incorporated temporal context that reflects physics and dynamics to enhance the touching expert model. We also fine-tuned the hovering expert model on data acquired on an inclined plane in order to benefit of height dimension variations in training. Our experiments on two datasets demonstrate that our mixture-of-experts outperforms the two main state-of-the-art methods.   

In addition, we propose the KIHT-public dataset, a database composed of more than 4k examples. Evaluations have consistently revealed similar conclusions between the KIHT-private dataset and the KIHT-public one, validating the relevance of this new public database. It enables researchers to rigorously test and refine algorithms, facilitating a deeper understanding of the complexities involved in handwriting trajectory reconstruction.

This work is dedicated to the handwriting reconstruction from data written on a tablet using the Digipen. The next challenge is to study the reconstruction of handwriting from data written on paper, which is noisier due to more friction. This task could be solved using transfer learning. In addition, as the Digipen can be used for learning to write in classroom, one should consider domain adaptation methods to transition from adult-acquired data to child-produced data. 

%% file: Sections/Acknowledgments.tex
\section{Acknowledgments}
\label{Acknowledgments}
This project is financed by the KIHT French-German bilateral ANR-21-FAI2-0007-01 project and these four partners, IRISA, KIT, Learn \& Go and Stabilo. This work was performed using HPC resources from GENCI-IDRIS (Grant 2021-AD011013148)

%% file: Sections/bio.tex
%\section*{Author Biographies}
%\label{Biographies}

\newpage

\textbf{Florent Imbert} is a Ph.D student at INSA. He hold a master's degree in applied mathematics from La Rochelle University. His Ph.D. is being carried out at IRISA and focuses on online handwriting trajectories reconstruction from kinematic sensors.

\textbf{Eric Anquetil} is a full professor at INSA Rennes. He leads the research on handwriting, gesture and drawing analysis and recognition at the IRISA laboratory. He is in charge of the "Innovation and Entrepreneurship" mission at INSA Rennes. His expertise is in human-machine interactivity through handwriting input and gesture commands.

\textbf{Yann Soullard} is an associate professor at Rennes 2 University (France) and a member of the IRISA computer science research institute. He holds a PhD degree from the Pierre and Marie Curie university of Paris, obtained in 2013. His research interests include deep learning, document analysis and handwriting recognition.

\textbf{Romain Tavenard} is a full professor at University of Rennes 2 since September 2022, following nine years as an assistant professor. He conducts research at IRISA, with a focus on machine learning and indexing for time series data, specifically for environmental time series.